\documentclass[journal, twoside, twocolumn]{IEEEtran}
\ifCLASSOPTIONcompsoc
  \usepackage[nocompress]{cite}

\else
  \usepackage{cite}
\fi
\ifCLASSINFOpdf
\else
\fi
\usepackage{amsmath}
\usepackage{caption}
\usepackage{amssymb}
\usepackage{amsmath}%新加的
\usepackage{balance}%新加的
\usepackage{algorithm}
\usepackage{algorithmic}
\usepackage{multirow}
\usepackage{subfigure}
\usepackage{graphicx}
\usepackage[table,xcdraw]{xcolor}
\usepackage[normalem]{ulem}
\usepackage{subfigure}
\usepackage{hyperref}
\useunder{\uline}{\ul}{}
\newtheorem{theorem}{Theorem}
\newtheorem{remark}{Remark}
\newtheorem{lemma}{Lemma}
\usepackage{ulem}
\usepackage{booktabs}
\usepackage{multirow}

\begin{document}
\title{Contrastive Continual Multi-view Clustering with Filtered Structural Fusion}

\author{~Xinhang~Wan,~Jiyuan~Liu,~Hao~Yu,~Ao~Li,~Xinwang~Liu$^{\dagger}$,~\IEEEmembership{Senior~Member,~IEEE},\\~Ke~Liang,~Zhibin~Dong,~En~Zhu$^{\dagger}$% <-this % stops a space
\IEEEcompsocitemizethanks{
\IEEEcompsocthanksitem X. Wan, J. Liu, H. Yu, X. Liu, K. Liang, Z. Dong, and E. Zhu are with School of Computer, National University of Defense Technology, Changsha, 410073, China. (E-mail: \{wanxinhang,\,liujiyuan13,\,yu\_haocs,\,xinwangliu,\,liangke22,\,dzb20,\,enzhu\} @nudt.edu.cn).
\IEEEcompsocthanksitem A. Li is with School of Computer Science and Technology, Harbin University of Science and Technology, 150080, China. (E-mail: ao.li@hrbust.edu.cn).
\IEEEcompsocthanksitem $^{\ast}$: Corresponding author.}
\thanks{Manuscript received Mar. 3, 2024.}
% <-this % stops an unwanted space
}

\markboth{IEEE Transactions on Neural Networks and Learning Systems}%
{WAN \MakeLowercase{\textit{et al.}}: Contrastive Continual Multi-view Clustering with Filtered Structural Fusion}
\maketitle

\begin{abstract}
Multi-view clustering thrives in applications where views are collected in advance by extracting consistent and complementary information among views. However, it overlooks scenarios where data views are collected sequentially, i.e., real-time data. Due to privacy issues or memory burden, previous views are not available with time in these situations. Some methods are proposed to handle it but are trapped in a stability-plasticity dilemma. In specific, these methods undergo a catastrophic forgetting of prior knowledge when a new view is attained. Such a catastrophic forgetting problem (CFP) would cause the consistent and complementary information hard to get and affect the clustering performance. To tackle this, we propose a novel method termed Contrastive Continual Multi-view Clustering with Filtered Structural Fusion (CCMVC-FSF). Precisely, considering that data correlations play a vital role in clustering and prior knowledge ought to guide the clustering process of a new view, we develop a data buffer with fixed size to store filtered structural information and utilize it to guide the generation of a robust partition matrix via contrastive learning. Furthermore, we theoretically connect CCMVC-FSF with semi-supervised learning and knowledge distillation. Extensive experiments exhibit the excellence of the proposed method.
\end{abstract}



\section{Introduction}\label{sec:introduction}

\IEEEPARstart{C}lustering aims to partition data samples into several groups according to their similarities and plays a crucial role in unsupervised learning \cite{8946880,8440680,9712412}. Existing clustering methods frequently assume that samples are collected from a single view and overlook multi-view applications \cite{9399655,8627941,8691702}. Data collected from multi-view is widespread in the world. For instance, a person can be described by appearance, social networks, characteristics, etc. How to label them is significant to some areas, such as psychology and biology.

Multi-view clustering, which leverages consistent and complementary information among views to perform clustering, has attracted broad attention due to its excellent clustering performance \cite{9093116,9646486,9718038,2020Multi}. To the best of our knowledge, existing methods can be roughly divided into five categories, i.e., multi-view subspace clustering (MVSC) 
\cite{8421595,8741173}, multiple kernel clustering (MKC) \cite{DBLP:journals/tkde/LiuLXLZWY22,DBLP:journals/pami/LiuLYLX23}, multi-view graph clustering (MVGC) \cite{zhan2018multiview,9769920,wen2023unpaired}, multi-view clustering based on matrix factorization (MVCMF) \cite{Wan_Liu_Liu_Wang_Wen_Liang_Zhu_Liu_Zhou_2023}, and deep multi-view clustering (DMVC) \cite{9839616,8999493}. In specific, MVSC assumes that data can represent themselves under a self-expressive framework and performs eigen-decomposition on the consensus coefficient matrix. For instance, by minimizing correlations among data objects from distinct subspaces, Wang et al. \cite{7160724} seek a robust subspace and conduct spectral clustering on it. MKC first constructs kernels of each view and integrates them as a consensus kernel, then conducts clustering on it. Utilizing the linear or nonlinear relationships among samples, MVGC first obtains graphs of each view and performs spectral clustering on a common graph. As pointed out by \cite{Gao2019MultiviewLM}, MVCMF factorizes the data features into two components, i.e., a consensus coefficient matrix and view-specific base matrices and the final result is attained on the consensus matrix. Different from the first four methods, DMVC follows a deep model and extracts more complex correlations among data \cite{9058997,10.1145/3474085.3475548,15198008220210920}. However, most of them fail to provide excellent clustering performance with theoretical support.
 \begin{figure*}[htbp]
 	\centering
 	
 	\subfigure{
 		\includegraphics[width=0.9\textwidth]{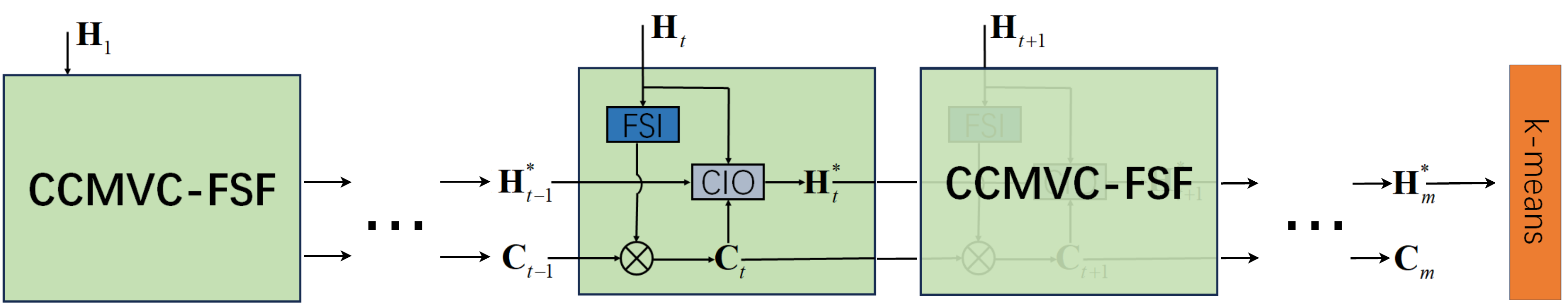}}
 	
 	\caption{ The basic framework of our proposed algorithm. The filtered structural information $\mathbf{C}_{t-1}$ of previous views is stored and updated in a buffer with fixed size. When the $t$-th view is collected, the model first extracts and updates the filtered structural information (FSI) to attain $\mathbf{C}_{t}$, then the consensus information optimization (CIO) is conducted with three parts of information, i.e., the partition matrix $\mathbf{H}_t$, the consensus matrix $\mathbf{H}^*_{t-1}$ of previous views, and the updated filtered structural information $\mathbf{C}_t$.
 	}
 	\label{alg_fig}
\end{figure*}

Despite existing methods achieving promising clustering performance, most are frequently under a static framework and neglect real-time data. Recent works assume that the views are collected in advance \cite{pmlr-v97-peng19a,10108535,9154578,9089264}. When a new view arrives, they need to integrate all views repeatedly, resulting in huge resource waste. Besides, in some practical situations, views cannot be held due to memory burden or privacy issues. For example, in a credit reporting system, a person's credit can be divided into several levels and identified by many factors, such as the consuming records on different banks. The level will be updated when a new bank is included in the system. However, previous information is hard to get owing to privacy considerations. Also, in a brain-computer interface system, the signal data at a moment contributes a new view, and the views are not able to be stored since the number is enormous.

To tackle the abovementioned problem, some researchers propose incremental multi-view clustering methods \cite{ZHOU201973, YIN2021260, 10.1145/3503161.3547864}. Zhou et al. \cite{ZHOU201973} store and update $m$ base kernels to leverage the previous information. By keeping a consensus similarity matrix, Yin et al. update it with the embedding of a coming view. Different from these methods with quadratic or cubic time and space complexity, the authors in \cite{10.1145/3503161.3547864} utilize a partition matrix to maintain the previous knowledge. Despite these methods enabling multi-view clustering to handle real-time data views from distinct perspectives, they all encounter a stability-plasticity dilemma. Specifically, these models pay much attention to the newly collected view and fail to leverage the information on prior knowledge adequately, and the knowledge of previous views will be gradually forgotten when new views are collected.
Consequently, consistent and complementary information among views is tough to attain, inevitably affecting the clustering performance. We refer to it as a catastrophic forgetting problem (CFP) in multi-view clustering. To the best of our knowledge, this problem has yet to be effectively solved so far since the storage and utilization of prior knowledge is challenging.

Given that in a clustering-induced task, the critical factor influencing the performance is the correlations among samples, especially the similar/dissimilar pairwise relationships. Some researchers focus on contrastive learning to attain data augmentations \cite{li2021contrastive, NEURIPS2021_10c66082, pmlr-v119-hassani20a,9852291,9577781}, which inspires a novel way to leverage structure information of views. However, most of them regard all of the samples except the data itself as negative samples, enlarging the data discrepancies in the same cluster. Some methods select $k$ nearest samples as positive samples and others as negative samples. It also inevitably broadens the distances of samples with the same label. Therefore, contrastive learning applied in clustering ought to consider positive/negative sample settings. Fortunately, in continual multi-view clustering, previous knowledge and clustering results could provide a reference to conduct contrastive learning and promote the generation of the partition matrix.

To this end,  we propose a novel method termed Contrastive Continual Multi-view Clustering with Filtered Structural Fusion (CCMVC-FSF), and the illustration of our framework is shown in Figure \ref{alg_fig}. We store the filtered structural information of previous views in a fixed-size data buffer. Once a new view arrives, a clustering then sample strategy will be conducted to select $m_p$ similar samples and $m_n$ different samples of each instance, and the information in the buffer will be updated. Then, the consensus knowledge will be generated by three parts: the clustering information of previous and the coming views, and the filtered structural information. In this way, contrastive learning is conducted in a more reasonable manner since previous knowledge is considered. We summarize the contributions of our work as follows,
 \begin{enumerate}
   \item We study a new paradigm on continual multi-view clustering, termed catastrophic forgetting problem (CFP). The previous views are not available with time in some scenarios due to memory burden or privacy considerations, it's tricky to extract consistent and complementary information among views.
 \item We propose a novel contrastive continual multi-view clustering method to overcome the CFP problem. A clustering then sample strategy is deployed to extract and update the filtered structure information of prior views, then the attained information will guide the clustering when a new view arrives.
\item We develop a two-step alternating optimization algorithm with proven convergence. Furthermore, comprehensive experiments exhibit the efficiency and effectiveness of our model.
\end{enumerate}

\section{Related Work}
\subsection{Multiple Kernel Clustering}
Let $\left\{\mathbf{x}_i\right\}_{i=1}^n \subseteq \mathcal{X}$ be a set of $n$ samples, and $\phi_p(\cdot):\mathbf{x} \in \mathcal{X} \mapsto \mathcal{H}_p$ be the $p$-th feature mapping function that projects $\mathbf{x}$ onto a reproducing kernel Hilbert space $\mathcal{H}_p(1 \leq p \leq m)$. In multiple kernel clustering scenarios, each sample can be denoted as $\phi_{\boldsymbol{\beta}}(\mathbf{x})=\left[\beta_1 \phi_1(\mathbf{x})^{\top}, \cdots, \beta_m \phi_m(\mathbf{x})^{\top}\right]^{\top}$, where $\boldsymbol{\beta}=\left[\beta_1, \cdots, \beta_m \right]^{\top}$ is the weight coefficient factor of $m$  base kernels $\left\{\kappa_p(\cdot, \cdot)\right\}_{p=1}^m$ \cite{10.5555/1953048.2021071,10.5555/2968826.2968972,9556554,9212617}. The consensus kernel can be constructed by the linear combinations of base kernels as follows,
 \begin{equation}
\begin{aligned}
 \kappa_{\boldsymbol{\beta}}\left(\mathbf{x}_i, \mathbf{x}_j\right)=\phi_{\boldsymbol{\beta}}\left(\mathbf{x}_i\right)^{\top} \phi_{\boldsymbol{\beta}}\left(\mathbf{x}_j\right)=\sum_{p=1}^m \beta_p^2 \kappa_p\left(\mathbf{x}_i, \mathbf{x}_j\right).
\end{aligned}
\end{equation}
Based on this, the clustering process can be formulated as,
\begin{equation}
\begin{aligned}\label{MKC}
\min _{\mathbf{H}, \boldsymbol{\beta}} & \operatorname{Tr}\left(\mathbf{K}_{\boldsymbol{\beta}}\left(\mathbf{I}_n-\mathbf{H H}^{\top}\right)\right) \\&\text { s.t. }  \mathbf{H} \in \mathbb{R}^{n \times k}, \mathbf{H}^{\top} \mathbf{H}=\mathbf{I}_k, \boldsymbol{\beta}^{\top} \mathbf{1}_m=1, \beta_p \geq 0, \forall p.
\end{aligned}
\end{equation}

The problem in Eq. \eqref{MKC} can be solved via an alternating optimization method in \cite{6031914}. However, the time complexity of MKC is  $\mathcal{O}\left(n^{3}\right)$ respecting sample number $n$, preventing it from being applied to large-scale scenes.
\subsection{Late Fusion Multi-view Clustering}
Late fusion multi-view clustering (LFMVC) has been recently proposed to reduce the high complexity of MKC \cite{ijcai2019p524, Liu2021LateFM,10011211,10.1145/3474085.3475204,8039105,10.1007/s11280-022-01012-7}. On the basis of the assumption that partition matrices among views share a consensus cluster structure, LFMVC seeks a common partition matrix by combining them with a linear transformation \cite{5459169,9190246,JOUIROU2019308}. It first outperforms eigen-decomposition on the base kernels $\{\mathbf{K}_p\}_{p=1}^m$ to obtain base partition matrices $\{\mathbf{H}_p\}_{p=1}^m$. After that, the consensus partition matrix is attained via the following formula:
\begin{equation}
\begin{aligned}
& \max _{\mathbf{H},\left\{\mathbf{W}_p\right\}_{p=1}^m, \boldsymbol{\beta}} \operatorname{Tr}\left(\mathbf{H}^{\top} \sum_{p=1}^m \beta_p \mathbf{H}_p \mathbf{W}_p\right)  \\
&\text { s.t. } \mathbf{H}^{\top} \mathbf{H}=\mathbf{I}_k, \mathbf{W}_p^{\top} \mathbf{W}_p=\mathbf{I}_k, 
 \sum_{p=1}^m \beta_p^2=1, \beta_p \geq 0, \forall p,
\end{aligned}
\end{equation}
where $\beta_p$ denotes the weight of $p$-th partition matrix, and $\mathbf{W}_p$ is a rotation matrix to match $\mathbf{H}_p$ with $\mathbf{H}$.

The optimization process can be solved  via an alternative algorithm  \cite{10011211}. Owing to the linear complexity respecting the sample number, LFMVC is capable of handling large-scale data. Despite extending MKC in some areas, it neglects situations where views increase with time, i.e., real-time data. When a new view arrives, it needs to store and recompute previous views, thus suffering from high memory and time burdens.

\subsection{Continual Multi-view Clustering}
Continual multi-view clustering concentrates on the field where views arrive sequentially \cite{10026925}. Given privacy issues or limited memory, previous views are avoided from being stored in raw format. Thus, the critical matter is to make use of the prior knowledge efficiently and effectively. Under the framework of multi-view graph clustering, Yin et al. \cite{YIN2021260} maintain a consensus similarity matrix to preserve previous information by the following objective:
\begin{equation}\label{SCGL}
\begin{aligned}
& \min _{\tilde{\mathbf{W}}_{t+1}} \mu_1\left\|\tilde{\mathbf{W}}_{t+1}-\mathbf{F}_t \mathbf{F}_t^T\right\|_F^2+\mu_2\left\|\tilde{\mathbf{W}}_{t+1}-\mathbf{F}^{(t+1)} \mathbf{F}^{(t+1)^T}\right\|_F^2 \\
&+\lambda\left\|\tilde{\mathbf{W}}_{t+1}\right\|_1,
\end{aligned}
\end{equation}
where $\tilde{\mathbf{W}}_{t+1}$ is the consensus similarity matrix, $\mathbf{F}_t$ denotes the spectral embedding of prior consensus similarity $\tilde{\mathbf{W}}_{t}$, $\mathbf{F}^{(t+1)}$ is the spectral embedding of the $(t+1)$-th view. However, the storage and update of $\tilde{\mathbf{W}}$ is $\mathcal{O}\left(n^{2}\right)$.

To alleviate the time burden of it, Wan et al.\cite{10.1145/3503161.3547864} propose a continual multi-view clustering method based on late fusion, which is formulated as,
\begin{equation}\label{final_IV_LFMVC}
\begin{aligned}
&\max _{\widetilde{\mathbf{H}}_{t},\mathbf{W}_{t}}  \operatorname{Tr}\left(\widetilde{\mathbf{H}}_{t}^{\top} \mathbf{H}_{t} \mathbf{W}_{t}\right)+
\lambda\operatorname{Tr}\left(\widetilde{\mathbf{H}}_{t}^{\top}\widetilde{\mathbf{H}}^*_{t-1}\right),\\
&\text { s.t. }  \widetilde{\mathbf{H}}_{t}^{\top} \widetilde{\mathbf{H}}_{t}=\mathbf{I}_{k}, \mathbf{W}_{t}^{\top} \mathbf{W}_{t}=\mathbf{I}_{k},
\end{aligned}
\end{equation}
where $\widetilde{\mathbf{H}}^*_{t-1}$ denotes the consensus partition matrix of previous views, and $\widetilde{\mathbf{H}}_{t}$ is a temporary component to match $\widetilde{\mathbf{H}}^*_{t-1}$ with ${\mathbf{H}}_{t}$.

Despite existing continual multi-view clustering being qualified for incremental views, all of them are faced with a stability-plasticity predicament. In specific, they pay much attention to the newly collected view and undergo a catastrophic forgetting of prior knowledge. For instance, in Eq. \eqref{SCGL} and \eqref{final_IV_LFMVC}, it is observed that the consensus information of previous views will gradually fade when new views arrive. Consequently, extracting consistent and complementary information among views is challenging to achieve, which harms the clustering results. To tackle it, we introduce a novel method termed Contrastive Continual Multi-view Clustering with Filtered Structural Fusion in the next section.

\section{Contrastive Continual Multi-view Clustering with Filtered Structural Fusion}
In this section, we first describe the motivation of our proposed method and then introduce its objective formula. After that, an alternating algorithm is provided to solve the resultant problem.
\subsection{The Design of A Fixed Data Buffer}
All existing multi-view clustering with continual data has trouble maintaining the abundant information of previous views. The reason is that they focus on the newly collected view and fail to utilize the knowledge of prior views sufficiently. Given that the correlations among samples play an essential role in clustering, we develop a fixed data buffer to store the correlations of samples in each view. However, storing a similarity matrix directly is redundant, and the similarity matrix is sensitive to the views with poor quality. Meanwhile, the integration of graphs among views results in enormous computation resources. Distinct from recent research \cite{ZHOU201973,YIN2021260}, we store the most $m_p$ similar samples and $m_n$ different samples of each instance in a buffer as positive/negative samples. The buffer will be updated based on the positive/negative settings when a new view is collected. It is worth mentioning that if one sample is considered a positive sample of previous views and viewed as a negative sample in the fresh view (small probability), it will be regarded as unrelated to the sample for simplicity. 
% Then the consensus partition matrix is generated with contrastive learning, since similar samples in previous views tend to share a close clustering partition, while different samples are inclined to share a distant clustering partition as prior information. In this way, the filtered structure information could guide the generation of the partition matrix as a contrastive term. Compared with existing multi-view clustering methods with contrastive learning, our proposed positive/negative sample settings are more reasonable since previous information could be listed as a prior.
\begin{remark}\label{buffer}
The size of the data buffer is $\min\left(n^2, \left(m_p+m_n\right)vn\right)$, where $v$ denotes the number of views has been collected so far. Therefore, we can fix it in a buffer with size $n^2$.
\end{remark}

In Remark \ref{buffer}, we analyze the maximum size of the data buffer. However, given that the number of positive/negative samples is not large, and most positive/negative samples are consistent among views, the buffer used to store them is sparse in practical.

\subsection{Clustering then Sample}
 Given a sample set $S=\left\{x_i\right\}_{i=1}^n$, we aim to select $m_p$ most similar samples as positive samples and $m_n$ most different samples as negative samples from $S$ for each instance. Consequently, the computation of positive/negative samples takes $\mathcal{O}\left(n^{2}\right)$ time complexity respecting sample number $n$, which is unsuitable for large-scale scenes. In light of this, we propose a sampling method with linear complexity to extend it to large-scale scenarios. In specific, we randomly select $r$ points $\tilde{S}_i=\left\{y_j\right\}_{j=1}^r$ from $S$ without replacement, then search for positive/negative samples of point $x_i$ from $\tilde{S}_i$. The sampling strategy is rational owing to the following theorem.
\begin{theorem}\label{ave_noeq}
Assume that $S=\left(x_1, \ldots, x_n\right)$ be a finite population with $n$ points, and $\tilde{S}_i=\left\{y_j\right\}_{j=1}^r$ denote a set randomly selected from $S$ without replacement, then we derive that
% \begin{equation}
%     a=\min _{1 \leq i \leq n} x_i \quad \text { and } \quad b=\max _{1 \leq i \leq n} x_i.
% \end{equation}

% Then for all $\epsilon \geq0$,
\begin{equation}
   % \operatorname{Pr}\left(\left|\frac{1}{r} \sum_{i=1}^r y_i-\frac{1}{n} \sum_{i=1}^n x_i\right| \geq t\right)  \precsim \sqrt{\left(\frac{1}{r}-\frac{1}{n}\right) \log \left(\frac{2}{\delta}\right)}
   \left|\frac{1}{r} \sum_{i=1}^r y_i-\frac{1}{n} \sum_{i=1}^n x_i\right|   \precsim \sqrt{\left(\frac{1}{r}-\frac{1}{n}\right) \log \left(\frac{2}{\delta}\right)}
\end{equation}
holds with probability at least $1-\delta$. The proof is given in the appendix.
\end{theorem}
\begin{theorem}
Based on the sampling strategy of Theorem \ref{ave_noeq}, and we denote $\sigma^2$, $\widehat{\sigma}^2$ as the variances of $S$ and $\tilde{S}_i$, respectively. The following concentration inequality holds with probability at least $1-2\delta$.
\begin{equation}
% \operatorname{Pr}\left(\left|\sigma - \widehat{\sigma}_n\right|\geq3(b-a) \sqrt{\frac{\log (3 / \delta)}{2 n}}\right) \leq \delta
\left|\sigma - \widehat{\sigma}\right|\precsim \sqrt{\frac{\log (3 / \delta)}{2 r}}
\end{equation}
The detailed proof is given in the appendix.
% \begin{equation}
%     a=\min _{1 \leq i \leq n} x_i \quad \text { and } \quad b=\max _{1 \leq i \leq n} x_i.
% \end{equation}
\end{theorem}
\begin{remark}\label{r_value}
 According to the above theorems, we know that the distribution of $\tilde{S}_i$ can be a good appropriation of $S$ when the selected number $r$ is large. However, larger $r$ makes the above bound tighter but costs more time. In our method, we let $r=\mathcal{O}\left(\sqrt{n}\right)$ to balance them. 
\end{remark}

An improved sampling strategy is developed to further eliminate the randomness in sampling. Specifically, when a new view is collected, and $\mathbf{H}_t$ is the corresponding partition matrix, we first conduct $k$-means on it to attain a cluster partition. Then, we seek positive samples from the points sampled in the same cluster and search for negative samples from instances sampled in different clusters. The reason that we do not directly assign positive samples with all samples in the same cluster is that the clustering process is initialization-sensitive and might generate wrong labels for some points, which will inevitably affect contrastive learning, cutting down the clustering performance subsequently. The proper contrastive settings in our method enable us to obtain a high-quality partition matrix. The framework of our clustering then sample strategy is summarized in Algorithm \ref{cluster_then_sample}.
\begin{algorithm}
	\renewcommand{\algorithmicrequire}{\textbf{Input:}}
	\renewcommand{\algorithmicensure}{\textbf{Output:}}
	\caption{The clustering then sample strategy.}
	\begin{algorithmic}[1]
		\REQUIRE $\mathbf{H}_t$, $\mathbf{W}_{t-1}$, $r$, $k$, $m_p$, $m_n$.
		\ENSURE $\mathbf{W}_{t}$.
		\STATE Conduct $k$-means with the information matrix $\mathbf{H}_t$. 
		\FOR{$i=1$ to $n$}
		\STATE Randomly select $r$ samples from the same (different) cluster(s) as sample $i$ to obtain $S_i^s$ ($S_i^d$).
		\STATE Seek the $m_p$ most similar samples from $S_i^s$ as positive samples, and seek the $m_n$ most different samples from $S_i^d$ as negative samples for the $i$-th sample.
		\ENDFOR
       \STATE Obtain the indicator matrix of $t$-th view $\tilde{\mathbf{W}}_t$ based on Eq. \eqref{pos_neg}.
       \STATE Update the filtered structural information $\mathbf{W}_{t-1}$ with $\tilde{\mathbf{W}}_t$ to obtain $\mathbf{W}_{t}$.
		%\STATE \textbf{return} $\mathbf{H}^*_m$
	\end{algorithmic}  
	\label{cluster_then_sample}
\end{algorithm}

\subsection{Contrastive Loss}
In this section, we focus on how filtered information guides the generation of the consensus matrix when a new view arrives. Given a representation matrix $\mathbf{H} \in \mathbb{R}^{n \times k}$, its similarity matrix can be obtained as follows,
\begin{equation}\label{con_loss}
    \begin{aligned} \mathbf{C} & =\frac{\mathbf{H H}^{ \top}}{\|\mathbf{H}\|_F^2} =\left[\begin{array}{cccc}C_{1,1} & C_{1,2} & \cdots & C_{1, n} \\ C_{2,1} & C_{2,2} & \cdots & C_{2, n} \\ \vdots & \vdots & \ddots & \vdots \\ C_{n, 1} & C_{n, 2} & \cdots & C_{n, n}\end{array}\right].\end{aligned}
\end{equation}

Based on this, the contrastive loss can be defined as,
\begin{equation}
\varrho\left(\mathbf{H}\right)=\operatorname{Tr}(\mathbf{C W}),
\end{equation}
where $\mathbf{W}$ is a pre-defined indicator matrix with three values as,
\begin{equation}\label{pos_neg}
W_{i, j}=\left\{\begin{array}{cl}
w_p & \text { if } \mathbf{h}_i \text { and } \mathbf{h}_j \text { are positive pairs, } \\
-w_n & \text { if } \mathbf{h}_i \text { and } \mathbf{h}_j \text { are negative pairs, }\\
0 & \text { otherwise. }
\end{array}\right.
\end{equation}

% Considering that the partition matrix satisfies that $\mathbf{H}^{\top} \mathbf{H}=\mathbf{I}_k$, the element of $\mathbf{H}$ is located in $\left[-1,1\right]$, 
We can bound the contrastive loss as the following Lemma.
\begin{lemma}
The contrastive loss of $\varrho\left(\mathbf{H}\right)$ exists in $\left[-\delta,\delta\right]$, where $\delta=\|\mathbf{W}\|_F$.

According to Cauchy-Schwarz inequality, it is obtained that:
\begin{equation}\label{CW_bound}
\begin{aligned}
  \left[ \operatorname{Tr}(\mathbf{C W}) \right]^2=\left(\sum_{i, j=1}^n w_{i, j} c_{i, j}\right)^2 \leq \sum_{i, j=1}^n w_{i, j}^2 \sum_{i, j=1}^n c_{i, j}^2 \\
  \leq\sum_{i, j=1}^n w_{i, j}^2= \|\mathbf{W}\|_F^2.
\end{aligned}
\end{equation}

\end{lemma}

\begin{remark}
As we analyzed before, $\mathbf{W}$ is a sparse matrix. Also, we can use the following strategy to accelerate the computation of Eq. \eqref{con_loss}: 

Given a data representation $\mathbf{H}$ in $l$ batches, where the batch size $b \ll n$, the contrastive loss can be approximated by:
\begin{equation}
\begin{aligned}
\varrho(\mathbf{H})=\sum_{i=1}^l \operatorname{Tr}\left(\mathbf{C}_i \mathbf{W}_i\right)
\end{aligned}
\end{equation}
where $\mathbf{C}_i$ and $\mathbf{W}_i$ are the similarity matrix and the pre-specified indicator matrix of the $i$-th batch sample pairs.
\end{remark}
\subsection{Formulation of Our Algorithm}
In the previous section, we focus on how to sufficiently utilize information from previous views to induce the generation of a consensus matrix when a new view arrives. In this section, we provide the formulation of our method.

Suppose that $t$-th view arrives and $\mathbf{H}_t$ denotes its partition matrix without prior guidance; the consensus partition matrix of previous views $\mathbf{H}^*_{t-1}$ has already been attained. We first impose a rotation matrix $\mathbf{M}_t$ on $\mathbf{H}_t$ to match $\mathbf{H}^*_{t-1}$. Then, the consensus matrix $\mathbf{H}^*_{t-1}$ will be updated by three parts of information, i.e., the partition matrix of the coming view, the consensus matrix, and the filtered structural information of prior knowledge. The formula is developed as follows,
\begin{equation}\label{final_CCMVC}
\begin{aligned}
\max _{\mathbf{H}^*_{t},\mathbf{M}_{t}} 
& \operatorname{Tr}\left({\mathbf{H}^*_{t}}^{\top} \left(\mathbf{H}^*_{t-1} +\mathbf{H}_{t}\mathbf{M}_{t}\right)\right)+
\lambda\operatorname{Tr}\left(\mathbf{C}_{t}\mathbf{W}_{t}\right)\\
&\text { s.t. }  {\mathbf{H}^*_{t}}^{\top} \mathbf{H}^*_{t}=\mathbf{I}_{k}, \mathbf{M}_{t}^{\top} \mathbf{M}_{t}=\mathbf{I}_{k},
\end{aligned}
\end{equation}
where
\begin{equation}
    \begin{aligned} \mathbf{C}_t & =\frac{\mathbf{H}^*_{t}{\mathbf{H}^*_{t}}^{\top}}{\|\mathbf{H}^*_{t}\|_F^2},\end{aligned}
\end{equation}
where $\lambda$ is a regularization term to balance the weight between partition matrix fusion and contrastive loss.

In this way, the consensus matrix $\mathbf{H}^*_{t}$ will guide the match of $\mathbf{H}_{t}$ and $\mathbf{H}^*_{t-1}$. As feedback, the matched matrices promote the quality of
$\mathbf{H}^*_{t}$. Also, prior knowledge is adequately used by the filtered structural information $\mathbf{W}_t$.

\subsection{Optimization}
We develop a two-step alternating optimization algorithm to solve the resultant problem in Eq. \eqref{final_CCMVC}. In each step, we optimize one variable while the other remains fixed.
\subsubsection{\texorpdfstring{${\mathbf{H}^*_{t}}$}--Subproblem}
Fixing $\mathbf{M}_{t}$, the formula in Eq. \eqref{final_CCMVC} can be reduced to:
\begin{equation}\label{opt_H}
\begin{aligned}
\max _{\mathbf{H}^*_{t}} 
& \operatorname{Tr}\left({\mathbf{H}^*_{t}}^{\top} \mathbf{A}\right)+
\widetilde\lambda\operatorname{Tr}\left({\mathbf{H}^*_{t}\mathbf{H}^*_{t}}^{\top} \mathbf{W}_{t}\right)
\text { s.t. } {\mathbf{H}^*_{t}}^{\top} \mathbf{H}^*_{t}=\mathbf{I}_{k},
\end{aligned}
\end{equation}
where $\mathbf{A}=\mathbf{H}^*_{t-1} +\mathbf{H}_{t}\mathbf{M}_{t}$, $\widetilde\lambda=\lambda/\|\mathbf{H}^*_{t}\|_F^2.$

It is obtained that it is a quadratic problem on the Stiefel manifold, according to \cite{nie_inf}, it can be efficiently solved via Algorithm \ref{algo_opt}.
\begin{algorithm}
	\renewcommand{\algorithmicrequire}{\textbf{Input:}}
	\renewcommand{\algorithmicensure}{\textbf{Output:}}
	\caption{The optimization of ${\mathbf{H}^*_{t}}$}
	\begin{algorithmic}[1]
		\REQUIRE $\mathbf{A}$, $\widetilde\lambda$, $\mathbf{W}_t$.
		\ENSURE $\mathbf{H}^*_t$.
		\STATE Initialize $\mathbf{H}^*_{t}$ satisfying that ${\mathbf{H}^*_{t}}^{\top} \mathbf{H}^*_{t}=\mathbf{I}_{k}$ randomly. 
		\WHILE {not converged}
		\STATE Update $\mathbf{K}=2\widetilde\lambda\mathbf{W}_{t}\mathbf{H}^*_{t}+\mathbf{A}$.
		\STATE Conduct singular value decomposition (SVD) on $\mathbf{K}$ to obtain $\mathbf{U} \in \mathbb{R}^{n \times k}$, $\mathbf{V} \in \mathbb{R}^{k \times k}$, where $\mathbf{K}=\mathbf{U} \boldsymbol{\Sigma} \mathbf{V}^{\top}$.
		\STATE Calculate $\mathbf{H}^*_{t}=\mathbf{U}\mathbf{V}^{\top}$.
		\ENDWHILE
		\
		%\STATE \textbf{return} $\mathbf{H}^*_m$
	\end{algorithmic}  
	\label{algo_opt}
\end{algorithm}
% \begin{enumerate}
% \item Initialize $\mathbf{H}^*_{t}$ satisfying that ${\mathbf{H}^*_{t}}^{\top} \mathbf{H}^*_{t}=\mathbf{I}_{k}$ randomly;
% \item Update $\mathbf{K}=2\widetilde\lambda\mathbf{W}_{t}\mathbf{H}^*_{t}+\mathbf{A}$;
% \item Conduct singular value decomposition (SVD) on $\mathbf{K}$ to obtain $\mathbf{U} \in \mathbb{R}^{n \times k}$, $\mathbf{V} \in \mathbb{R}^{k \times k}$, where $\mathbf{K}=\mathbf{U} \boldsymbol{\Sigma} \mathbf{V}^{\top}$;
% \item Calculate $\mathbf{H}^*_{t}=\mathbf{U}\mathbf{V}^{\top}$;
% \item Iteratively perform 2-4 until convergence.
% \end{enumerate}

\subsubsection{\texorpdfstring{${\mathbf{M}_{t}}$}--Subproblem}
Fixing $\mathbf{H}^*_{t}$, the formula in Eq. \eqref{final_CCMVC} can be rewritten as,
\begin{equation}\label{opt_M}
\begin{aligned}
\max _{\mathbf{M}_{t}} 
& \operatorname{Tr}\left({\mathbf{M}_{t}}^{\top} \mathbf{B}\right)
\text { s.t. }  \mathbf{M}_{t}^{\top} \mathbf{M}_{t}=\mathbf{I}_{k},
\end{aligned}
\end{equation}
where $\mathbf{B}={\mathbf{H}^{t}}^{\top} \mathbf{H}^*_{t}$.

\begin{remark}
Suppose that matrix $\mathbf{B}$ has the singular value decomposition (SVD) form as $\mathbf{B}=\mathbf{S} \boldsymbol{\Sigma} \mathbf{V}^{\top}$, the optimization problem in Eq. (\ref{opt_M}) can be solved by a closed-form solution in \cite{ijcai2019-524} as $\mathbf{M}_t=\mathbf{S}  \mathbf{V}^{\top}$. And the time complexity of solving it is linear complexity about the sample number. The detailed proof is provided as follows.
\end{remark}
\begin{theorem}
Suppose that matrix $\mathbf{B}$ has the singular value decomposition (SVD) form as $\mathbf{B}=\mathbf{S} \boldsymbol{\Sigma} \mathbf{V}^{\top}$, Eq. \eqref{opt_M} can be rewritten as,
\begin{equation}
\begin{aligned}
\operatorname{Tr}\left({\mathbf{M}_{t}}^{\top} \mathbf{B}\right)=\operatorname{Tr}\left({\mathbf{M}_{t}}^{\top} \mathbf{S} \boldsymbol{\Sigma} \mathbf{V}^{\top}\right).
\end{aligned}
\end{equation}
Let $\mathbf{Q}=\mathbf{V}^{\top}{\mathbf{M}_{t}}^{\top} \mathbf{S}$, we have  $\mathbf{Q}\mathbf{Q}^{\top}=\mathbf{I}_{k}$. Hence, we can take $\operatorname{Tr}\left(\mathbf{V}^{\top}{\mathbf{M}_{t}}^{\top} \mathbf{S}\boldsymbol{\Sigma}\right)=\operatorname{Tr}\left(\mathbf{Q}\boldsymbol{\Sigma}\right)\leq \sum_{i=1}^k \sigma_i$. Therefore, to maximize the objective value in Eq. \eqref{opt_M}, $\mathbf{M}_t=\mathbf{S}  \mathbf{V}^{\top}$.    
\end{theorem}

Assume that the total view number is $m$, after obtaining the consensus partition matrix $\mathbf{H}_m^*$, we conduct $k$-means on it to attain the final clustering results. The whole optimization process of our proposed method is summarized in Algorithm \ref{algo_whole}.
\begin{algorithm}
	\renewcommand{\algorithmicrequire}{\textbf{Input:}}
	\renewcommand{\algorithmicensure}{\textbf{Output:}}
	\caption{Contrastive Continual Multi-view Clustering with Filtered Structural Fusion (CCMVC-FSF)}
	\label{alg:1}
	\begin{algorithmic}[1]
		\REQUIRE $\left\{\mathbf{H}_t\right\}_{t=1}^{m}$, $k$, $\lambda$ and $\varepsilon_0$.
		\ENSURE $\mathbf{H}^*_m$.
		\STATE Initialize $\mathbf{H}^*_0=\mathbf{H}_1$, $\mathbf{W}_0=\mathbf{0}$. 
		\FOR{$t=1$ to $m$}
		\STATE Update the filtered structure information $\mathbf{W}_t$.
		\STATE Initialize $\mathbf{M}_{t}=\mathbf{I}_{k}$ and $i=1$.
		\WHILE {not converged}
		\STATE Update $\mathbf{H}_{t}^*$ by solving Eq. (\ref{opt_H}).
		\STATE Update $\mathbf{M}_{t}$ by solving Eq. (\ref{opt_M}).
		\STATE $i\gets i+1$.
		\ENDWHILE{$\left(obj^i-obj^{i-1}\right)/obj^i\le\varepsilon_0$}
		\ENDFOR
		%\STATE \textbf{return} $\mathbf{H}^*_m$
	\end{algorithmic}  
	\label{algo_whole}
\end{algorithm}
\subsection{Convergence Analysis}\label{convergence_section}
Most MVC methods, such as \cite{pan2021multi}, fail to be proven to converge, while the convergence of CCMVC-FSF is theoretically guaranteed. Based on Cauchy-Schwartz inequality, we have
\begin{equation}\label{cauchy}
\begin{aligned}
&\operatorname{Tr}\left({\mathbf{H}^*_{t}}^{\top} \left(\mathbf{H}^*_{t-1} +\mathbf{H}_{t}\mathbf{M}_{t}\right)\right)
% = \operatorname{Tr}\left({\mathbf{H}^*_{t}}^{\top} \mathbf{H}^*_{t-1}\right) +\operatorname{Tr}\left({\mathbf{H}^*_{t}}^{\top}\mathbf{H}_{t}\mathbf{M}_{t}\right)\\ &
\leq \Vert \mathbf{H}^*_{t} \Vert_F \Vert\mathbf{H}^*_{t-1} \Vert_F\\
&+\Vert \mathbf{H}^*_{t} \Vert_F \Vert\mathbf{H}_{t} \mathbf{M}_{t} \Vert_F=2k.
\end{aligned}
\end{equation}
Combining Eq. \eqref{CW_bound} with \eqref{cauchy}, we have
\begin{equation}\label{upper_bound}
\begin{aligned}
\operatorname{Tr}\left({\mathbf{H}^*_{t}}^{\top} \left(\mathbf{H}^*_{t-1} +\mathbf{H}_{t}\mathbf{M}_{t}\right)\right)+
\lambda\operatorname{Tr}\left(\mathbf{C}_{t}\mathbf{W}_{t}\right)\leq \\ 2k+\lambda\|\mathbf{W}_t\|_F.
\end{aligned}
\end{equation}
Considering that $\|\mathbf{W}_t\|_F$ is a constant value, it is easy to obtain that the objective function of Eq. \eqref{final_CCMVC} has an upper bound. Then, we will prove that the objective value in Eq. \eqref{final_CCMVC} monotonically increases along with iterations. For ease of expression, we simplify the objective in Eq. \eqref{final_CCMVC} as,
\begin{equation}
\begin{aligned}
\min_{\mathbf{H}_{t}^*,\mathbf{M}_{t}} \mathcal{J}\left(\mathbf{H}_{t}^*,\mathbf{M}_{t}\right).
\end{aligned}
\end{equation}

It is obtained that the optimization process consists of two steps at each iteration, i.e., $\mathbf{H}_{t}^*$, and $\mathbf{M}_{t}$ subproblems. Let superscript $i$ denote the optimization process at  $i$-th round. We have:
\subsubsection{\texorpdfstring{${\mathbf{H}_{t}^*}$}- Subproblem} 
With $\mathbf{M}_{t}$ fixed, $\mathbf{H}_{t}^*$ can be updated via Eq. (\ref{opt_H}), resulting in:
\begin{equation}\label{H_ne}
\begin{aligned}
 \mathcal{J}\left({\mathbf{H}_{t}^*}^{\left(i+1\right)}, {\mathbf{M}_{t}}^{\left(i\right)}\right) \geq \mathcal{J}\left({\mathbf{H}_{t}^*}^{\left(i\right)}, {\mathbf{M}_{t}}^{\left(i\right)}\right).
\end{aligned}
\end{equation}
\subsubsection{\texorpdfstring{${\mathbf{M}_{t}}$}- Subproblem} 
With ${\mathbf{H}_{t}^*}$ fixed, $\mathbf{M}_{t}$ can be updated via Eq. (\ref{opt_M}), resulting in:
\begin{equation}\label{M_ne}
\begin{aligned}
\mathcal{J}\left({\mathbf{H}_{t}^*}^{\left(i+1\right)}, {\mathbf{M}_{t}}^{\left(i+1\right)}\right) \geq \mathcal{J}\left({\mathbf{H}_{t}^*}^{\left(i+1\right)}, {\mathbf{M}_{t}}^{\left(i\right)}\right).
\end{aligned}
\end{equation}

To sum up Eq.\eqref{H_ne} with Eq.\eqref{M_ne}, we have:
\begin{equation}\label{final_ne}
\begin{aligned}
\mathcal{J}\left({\mathbf{H}_{t}^*}^{\left(i+1\right)}, {\mathbf{M}_{t}}^{\left(i+1\right)}\right) \geq \mathcal{J}\left({\mathbf{H}_{t}^*}^{\left(i\right)}, {\mathbf{M}_{t}}^{\left(i\right)}\right),
\end{aligned}
\end{equation}
which indicates that the objective value monotonically increases along with iterations. So, our proposed method is convergent theoretically.
\section{Discussion}
\subsection{Connection with Semi-Supervised Learning}
Semi-supervised learning leverages unlabeled data when labels are limited or expensive to get \cite{NEURIPS2022_15dce910}. It is easy to observe that our proposed CCMVC-FSF can be used in semi-supervised learning when positive samples are points with the same label and negative samples are considered as points with different labels.
\subsection{Connection with Knowledge Distillation}
Knowledge distillation aims to transfer knowledge from a large model to a smaller model \cite{NEURIPS2018_019d385e}. In our framework, the filtered structure information can be regarded as a teacher to guide the generation of the consensus partition matrix. Also, continual contrastive clustering can be implemented under a deep learning framework with knowledge distillation, where clustering results (models) of previous views can be listed as a teacher of the model of the newly collected view.

\subsection{The Extension of Continual Learning}
Motivated by continual learning, we propose a method termed Contrastive Continual Multi-view Clustering with Filtered Structural Fusion to overcome the CFP problem in multi-view clustering. However, our work is not a simple imitation of it. The differences are listed as follows.

Most continual learning methods store historical data in a buffer to replay them when new data arrives \cite{10.5555/3504035.3504439,10.5555/3454287.3454319}. However, it results in two problems: privacy issues and the memory burden. When a new data stream comes, the sample number or the class number might change, and the previous data needs to be replayed to stabilize performance in previous tasks. When the sample number/distribution changes, the buffer size will be enlarged, or the sample proportion of previous tasks will decrease. On the contrary, in MVC, the views of data provide consistent and complementary information to conduct clustering. Since the sample number and class probability share among views, storing and updating the correlations is reasonable in continual multi-view clustering. Furthermore, the size of the data buffer can be fixed according to Remark \ref{buffer}.

% However, in most cases, storing the data itself is unable to achieve caused by the privacy considerations. Our proposed method that stores the relationships among data alleviates the problem. Meanwhile, our buffer is fixed since it isin MVC, the views of data provide consistent and complementary information to conduct clustering. Since the sample number and class probability share among views, storing and updating the correlations is reasonable in continual multi-view clustering. On the contrary, in most continual learning methods, the sample number or the class number might change, and the data need to be replayed to make the performance in previous tasks stable. When the sample number/distribution changes, the buffer size will be enlarged, or the sample proportion of previous tasks will decrease.

% The goals of sample selection are different. We select positive/negative samples to induce the generation of a partition matrix. Similar instances in previous views tend to share a close clustering partition when a new view arrives. While [1] aims to reduce the training cost of the data could and selects informative samples to train instead. Tasks are different. Our work aims to solve an unsupervised task, there is a lack of labels for samples, and the correlations among samples play a vital role in clustering. Therefore, the storage and update of them is reasonable. [1] focuses on a supervised task, the labeled data can provide more information to select ‘important’ samples and train the classifier.

\subsection{How to Handle Views with Different Dimensions}
In our paper, when a new view arrives, we first project the samples into a shared space via an eigendecomposition method \cite{pmlr-v139-liu21l}, and the dimension is equal to the class number. It has two reasons: the first reason is that when setting the dimension as $k$, the eigendecomposition on the data matrices leads to an information matrix $\mathbf{H} \in \mathbb{R}^{n \times k}$. The information matrix can be regarded as a soft partition matrix, which benefits the subsequent k-means to conduct the clustering then sample operation in our method. The second reason is that mapping data matrices of different views into a shared space simplifies the optimization process. Furthermore, we can also map data matrices into diverse spaces and fuse the information with different dimensions. For instance, \cite{Wan_Liu_Liu_Wang_Wen_Liang_Zhu_Liu_Zhou_2023} proposes a method to integrate matrices with different sizes, which inspires us to match matrices with different dimensions. Also, our method is general since the information matrices can be generated by other dimension reduction methods, such as encoder, matrix factorization, PCA, etc.
\subsection{The Specific Values of \texorpdfstring{${\mathbf{W}_{t}}$}-}
From Eq. \eqref{final_CCMVC}, it is seen that when the ratio of $w_p$ and $w_n$ is a constant value, the weight between partition matrix fusion and contrastive loss will be adjusted by $\lambda$. In our implementation, we set the ratio as 5. The weight to balance them is an interesting problem, and many contrastive learning methods are designed to investigate it \cite{NEURIPS2020_d89a66c7,9577669}. In our paper, we focus on overcoming CFP in multi-view clustering, and we might study how to set it more rational in our future work.
\begin{table}[htbp]
	\centering
 \caption{Datasets used in our experiments.} 
 % \label{dataset}
\begin{tabular}{cccc}
\toprule
Dataset  & Samples & Views & Clusters \\ \hline
3Sources & 169     & 3     & 6        \\
Mfeat    & 2000    & 12    & 10       \\
Citeseer & 3312    & 2     & 6        \\
SUNRGBD  & 10335   & 2     & 45       \\
Reuters  & 18756   & 5     & 6        \\
YTB10    & 38654   & 4     & 10       \\ \bottomrule
\label{dataset}
\end{tabular}
\end{table}

\begin{table*}[]
\centering
	\caption{Empirical evaluation and comparison of CCMVC-FSF with nine compared methods on six benchmark datasets in terms of ACC, NMI, and Purity. Note that '\--{}' indicates the method fails to run smoothly due to the out-of-memory error or other reasons. The best results are marked in bold, and the second best is underlined.}
	\label{cmp_result}
% \usepackage[normalem]{ulem}
\begin{tabular}{ccccccccccc}
\toprule
         & LMKKM & MFLVC       & SDMVC       & SCGL        & AE2-Nets & SFMC  & FMCNOF & IMSC        & CMVC           & CCMVC-FSF      \\ \hline
\multicolumn{11}{c}{ACC}                                                                                                               \\  \hline
3Sources & 34.91 & {\ul 66.27} & 63.91       & 62.78       & 29.23    & 34.91 & 62.09  & 54.44       & 62.31          & \textbf{69.23} \\
Mfeat    & 34.91 & 86.20       & 49.45       & 86.75       & 87.82    & 56.90 & 56.95  & {\ul 94.20} & 93.45          & \textbf{94.95} \\
Citeseer & 20.63 & 25.09       & 41.33       & {\ul 46.86} & 38.29    & 21.83 & 33.42  & 30.80       & 20.56          & \textbf{49.37} \\
SUNRGBD  & 18.46 & 13.22       & 19.22       & {\ul 19.86} & 11.82    & 11.02 & 19.67  & 15.06       & 18.37          & \textbf{19.94} \\
Reuters  & -   & -         & -         & 43.85       & 24.14    & 25.13 & 36.83  & 36.70       & {\ul 50.50}    & \textbf{51.65} \\
YTB10    & -   & 66.12       & 52.65       & -         & 70.75    & 55.80 & 43.42  & 52.92       & {\ul 86.19}    & \textbf{91.44} \\ \hline
\multicolumn{11}{c}{NMI}                                                                                                               \\ \hline
3Sources & 14.22 & 52.76       & {\ul 61.34} & 59.53       & 7.16     & 6.28  & 51.96  & 48.33       & 55.19          & \textbf{63.39} \\
Mfeat    & 14.22 & 85.53       & 51.90       & 84.26       & 80.18    & 68.15 & 55.47  & {\ul 88.43} & 86.82          & \textbf{89.47} \\
Citeseer & 1.64  & 4.08        & 19.00       & 21.51       & 16.77    & 6.49  & 12.68  & 10.34       & {\ul 24.33}    & \textbf{24.47} \\
SUNRGBD  & 21.21 & 13.69       & 14.11       & {\ul 24.42} & 17.62    & 2.30  & 15.66  & 19.59       & 23.37          & \textbf{24.93} \\
Reuters  & -   & -         & -         & 21.41       & 0.75     & 12.80 & 17.06  & 16.25       & {\ul 29.28}    & \textbf{30.33} \\
YTB10    & -   & 69.48       & 58.59       & -         & 73.17    & 77.46 & 39.15  & 55.16       & {\ul 85.74}    & \textbf{87.12} \\ \hline
\multicolumn{11}{c}{Purity}                                                                                                            \\ \hline
3Sources & 47.34 & 68.05       & {\ul 76.33} & 75.15       & 41.42    & 35.50 & 66.86  & 72.78       & 74.56          & \textbf{78.11} \\
Mfeat    & 47.34 & 86.20       & 49.85       & 86.75       & 87.82    & 57.40 & 57.10  & {\ul 94.20} & 93.45          & \textbf{94.95} \\
Citeseer & 24.12 & 27.87       & 46.26       & {\ul 48.61} & 40.92    & 24.67 & 33.67  & 32.76       & 46.53          & \textbf{50.91} \\
SUNRGBD  & 36.35 & 23.12       & 21.53       & {\ul 39.06} & 28.42    & 11.47 & 25.15  & 34.35       & \textbf{39.68} & 37.96          \\
Reuters  & -   & -         & -         & 50.85       & 31.29    & 35.40 & 40.42  & 41.99       & {\ul 53.14}    & \textbf{55.48} \\
YTB10    & -   & 71.11       & 55.41       & -         & 73.69    & 74.10 & 46.53  & 59.02       & {\ul 87.49}    & \textbf{91.44}\\ \bottomrule
\end{tabular}
\end{table*}
\section{Experiments}
In this section, we conduct experiments to verify the superiority of CCMVC-FSF on six datasets with several state-of-the-art multi-view clustering methods. 
\subsection{Setting}
We utilize six datasets ranging from 169 to 38654 instances in our experiments, including 3Sources\footnote{\url{http://mlg.ucd.ie/datasets/3sources.html}}, Mfeat\footnote{\url{https://archive.ics.uci.edu/dataset/72/multiple+features}}, Citeseer\footnote{\url{http://www.cs.umd.edu/projects/linqs/projects/lbc/}}, SUNRGBD\footnote{\url{http://rgbd.cs.princeton.edu/}}, Reuters\footnote{\url{http://kdd.ics.uci.edu/databases/reuters21578/}}, and YTB10 \footnote{\url{http://archive.ics.uci.edu/ml/datasets/YouTube+Multiview+Video+Games+Dataset}}. For all datasets, we assume that the number of clusters $k$ is provided and set it as the actual number of classes. Detailed information on these datasets is listed in Table \ref{dataset}.

Nine state-of-the-art multi-view clustering methods are introduced in our paper. They are summarized as follows:  
\begin{enumerate}
\item \textbf{LMKKM} \cite{Mehmet2014Localized}. Under the framework of multi-kernel clustering, LMKKM fuses the localized kernels as a consensus kernel to attain a better representation.
\item  \textbf{MFLVC} \cite{Xu_2022_CVPR}. It is a deep multi-view clustering method, which proposes a new framework of multi-level feature learning for MVC.
\item  \textbf{SDMVC} \cite{9839616}. SDMVC proposes a self-supervised discriminative feature learning for deep multi-view clustering.
\item \textbf{SCGL} \cite{ YIN2021260}. It proposes an efficient incremental multi-view spectral clustering method with sparse and connected graph learning.
\item \textbf{AE2-Nets} \cite{8953969}.  It integrates information from heterogeneous sources into an intact representation by utilizing the auto-encoder.
% \item \textbf{SMKKM} \cite{9857664}. SMKKM combines supervised kernel alignment criterion with multi-kernel clustering and uses a reduced gradient descent algorithm to optimize it.
\item \textbf{SFMC} \cite{9146384}. SFMC uses a bipartite graph to reduce the high complexity of multi-view graph clustering by presenting a scalable and parameter-free graph fusion framework.
\item \textbf{FMCNOF} \cite{9305974}. It is a fast multi-view clustering model via nonnegative and orthogonal factorization.
% \item \textbf{OPLFMVC} \cite{pmlr-v139-liu21l}. OPLFMVC is a batch of late fusion multi-view clustering. It proposes a one-pass method to attain a discrete partition matrix.
\item \textbf{IMSC} \cite{ZHOU201973}. IMSC handles multi-view graph clustering by integrating views one by one in an incremental way.

\item \textbf{CMVC} \cite{10.1145/3503161.3547864}. It deals with continual multi-view clustering by a late fusion method, and its time complexity is linear, respecting the sample number.
 \end{enumerate}
 \begin{figure*}[htbp]

 \centering
	% \subfigure{
	% 	\includegraphics[width=0.23\textwidth]{pdf/ar10p_iter.pdf}}
% \hspace{-0.2cm}
	\subfigure{
		\includegraphics[width=0.23\textwidth]{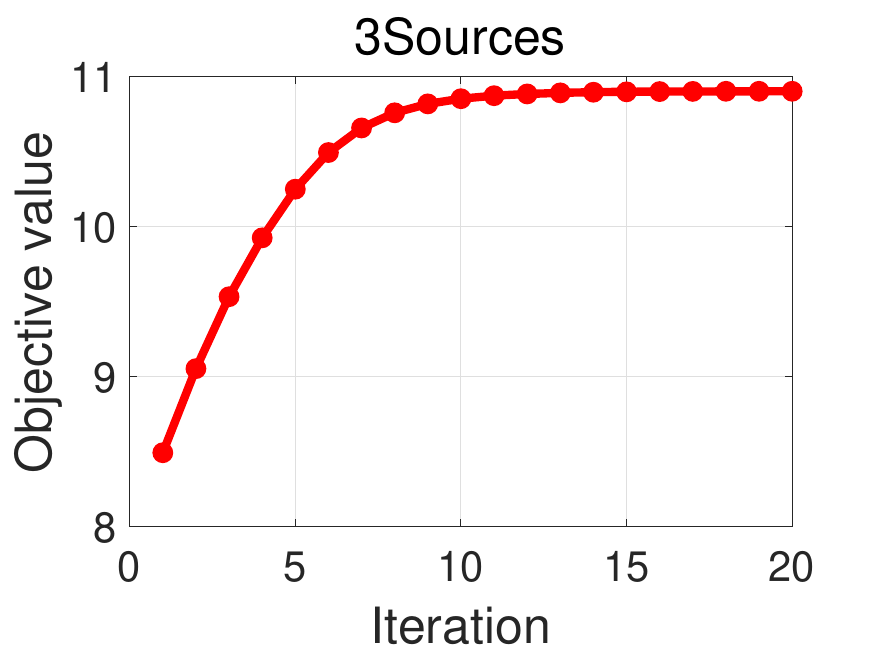}}
% \hspace{-0.2cm}
% 	\subfigure{
% 		\includegraphics[width=0.3\textwidth]{pdf/mfea.pdf}}
% % \hspace{-0.2cm}	
% \subfigure{
% 		\includegraphics[width=0.3\textwidth]{pdf/awa_iter.pdf}}
% \hspace{-0.2cm}
	\subfigure{
		\includegraphics[width=0.23\textwidth]{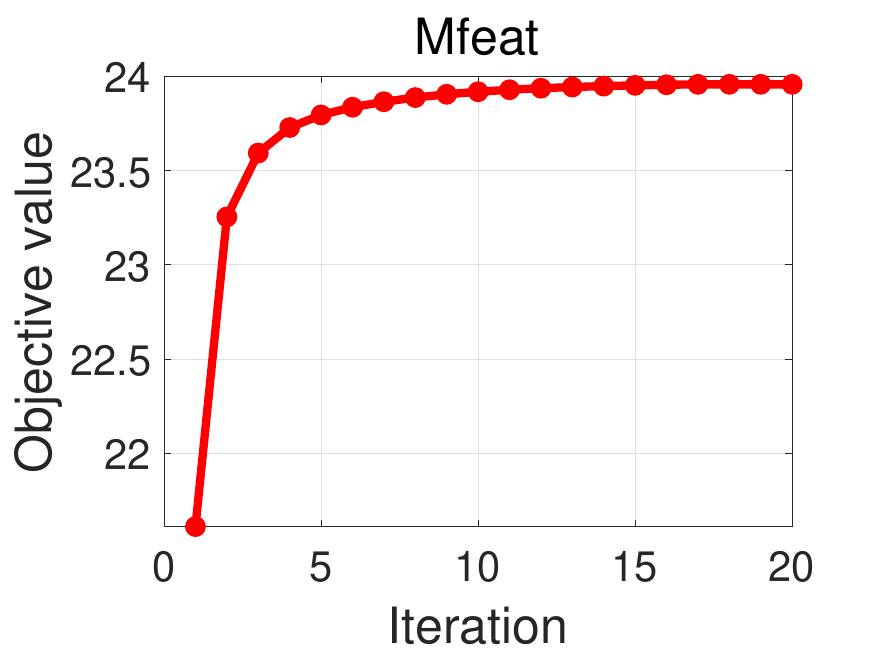}}
% \hspace{-0.2cm}
	\subfigure{
		\includegraphics[width=0.23\textwidth]{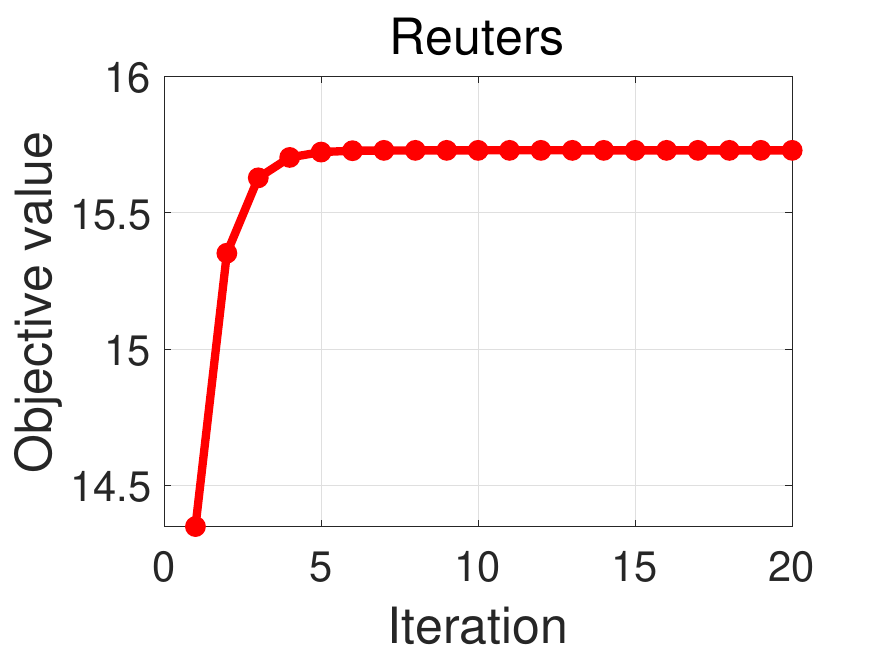}}
% \hspace{-0.2cm}	
\subfigure{
		\includegraphics[width=0.23\textwidth]{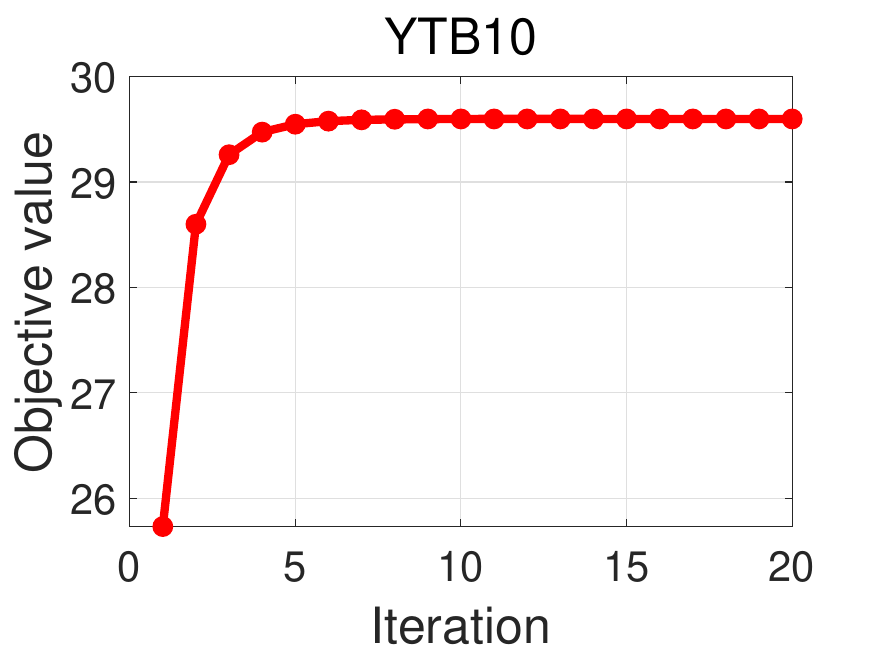}}
	%\subfigure{
	%	\includegraphics[width=0.3\textwidth]{pdf/olympics_iter.pdf}}
	
	\caption{The objective values of CCMVC-FSF vary with iterations on 3Sources, Mfeat, Reuters, and YTB10.}
\label{fig_obj}
\end{figure*}

\begin{figure*}[htbp]
	\centering

	\subfigure[1-st iteration]{
		\includegraphics[width=0.3\textwidth]{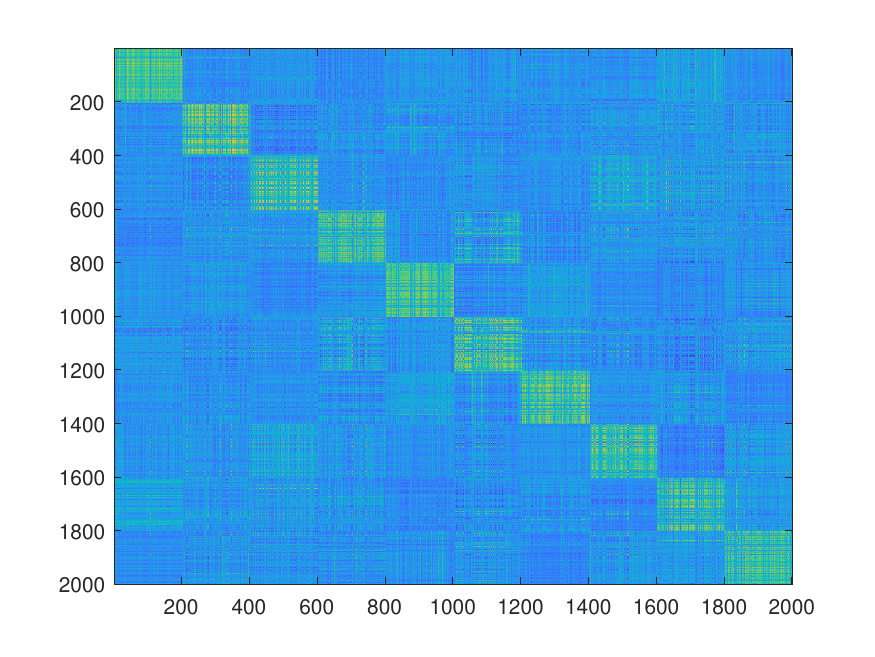}}
% \hspace{-0.2cm}
	% \subfigure[3-rd iteration]{
	% 	\includegraphics[width=0.23\textwidth]{pdf/mfeat_qua_iter3.pdf}}
% \hspace{-0.2cm}
% 	\subfigure{
% 		\includegraphics[width=0.3\textwidth]{pdf/mfea.pdf}}
% % \hspace{-0.2cm}	
% \subfigure{
% 		\includegraphics[width=0.3\textwidth]{pdf/awa_iter.pdf}}
% \hspace{-0.2cm}
	\subfigure[5-th iteration]{
		\includegraphics[width=0.3\textwidth]{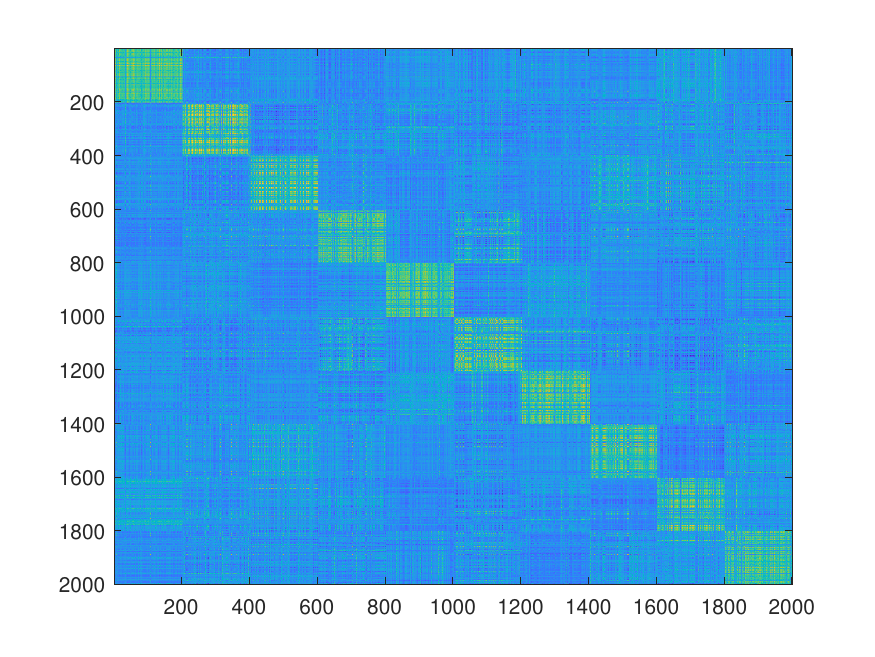}}
% \hspace{-0.2cm}
	\subfigure[10-th iteration]{
		\includegraphics[width=0.3\textwidth]{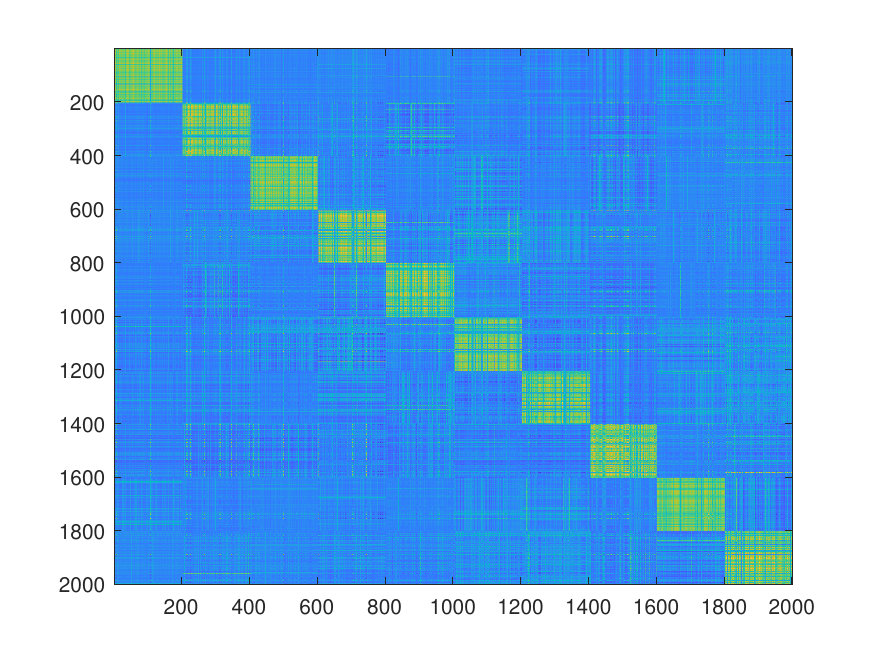}}
% \hspace{-0.2cm}	

	%\subfigure{
	%	\includegraphics[width=0.3\textwidth]{pdf/olympics_iter.pdf}}
	
\centering
	\caption{Visualization of the similarity matrix computed by  $\mathbf{H}^*_{t}{\mathbf{H}^*_{t}}^{\top}$ on Mfeat varies with iterations.}
\label{fig_matrix_qua}
\end{figure*}

 \begin{figure*}[htbp]
	\centering

% \hspace{-0.2cm}
	\subfigure{
		\includegraphics[width=0.23\textwidth]{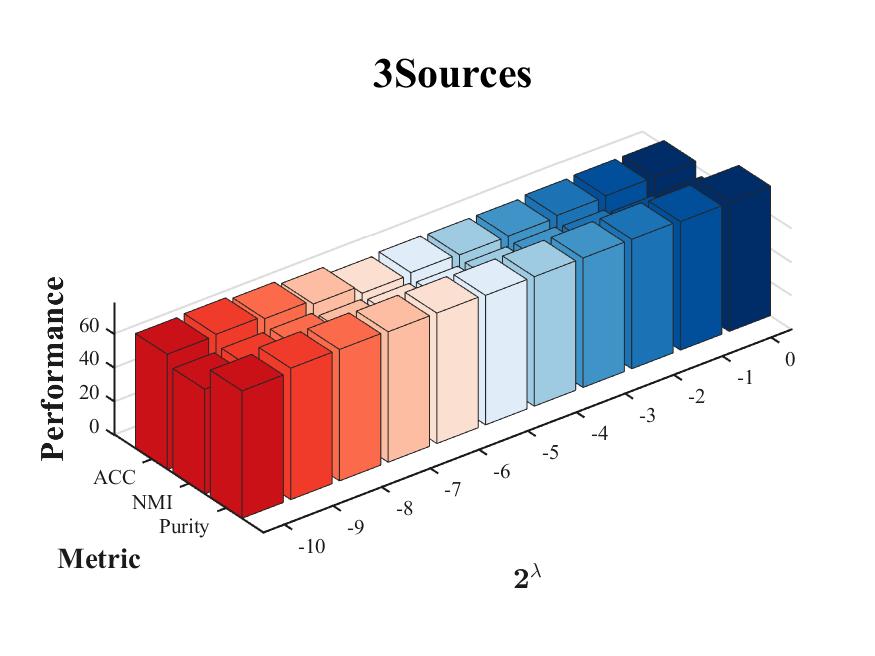}}
% \hspace{-0.2cm}
% 	\subfigure{
% 		\includegraphics[width=0.3\textwidth]{pdf/mfea.pdf}}
% % \hspace{-0.2cm}	
% \subfigure{
% 		\includegraphics[width=0.3\textwidth]{pdf/awa_iter.pdf}}
% \hspace{-0.2cm}
	\subfigure{
		\includegraphics[width=0.23\textwidth]{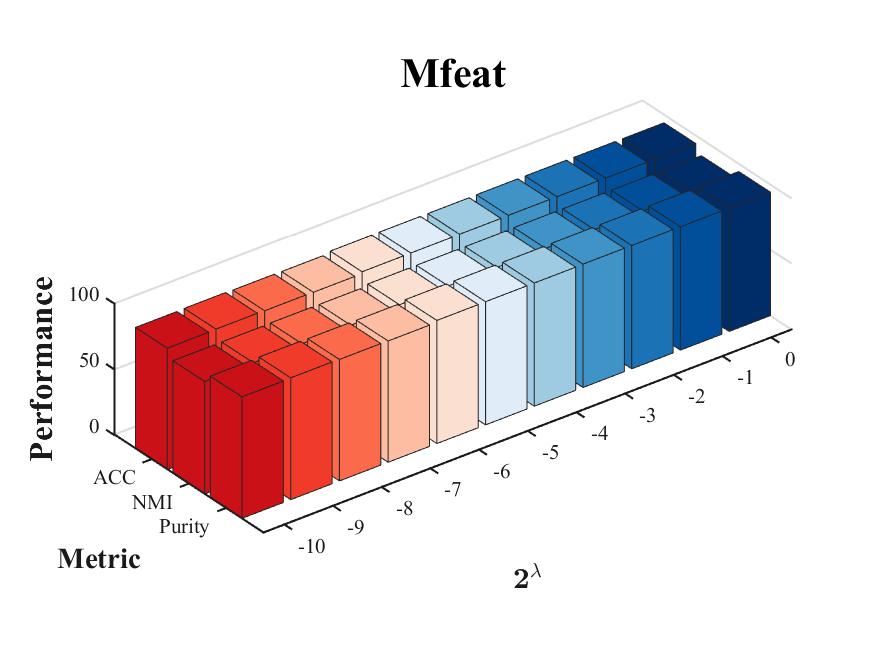}}
% \hspace{-0.2cm}
	\subfigure{
		\includegraphics[width=0.23\textwidth]{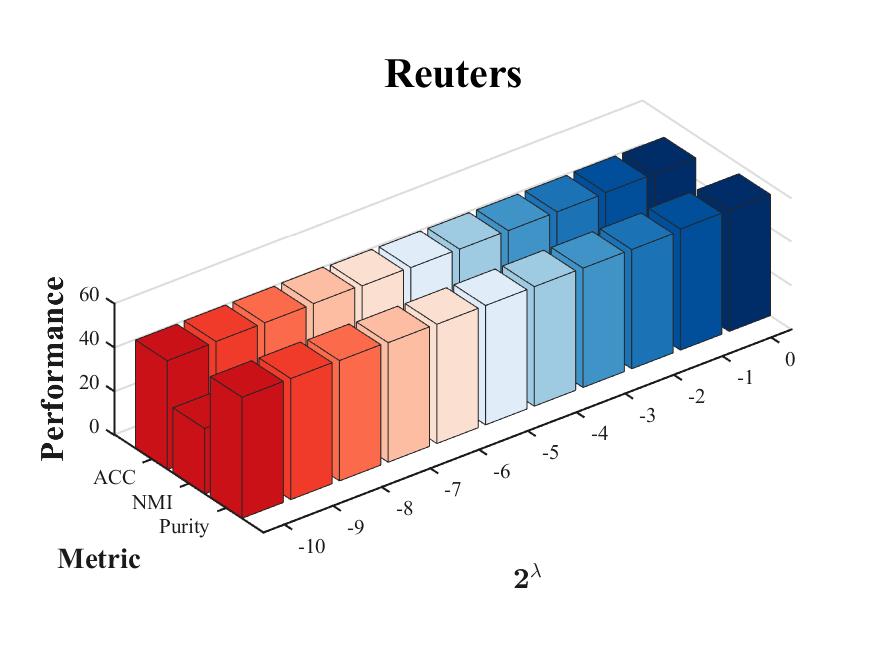}}
% \hspace{-0.2cm}	
\subfigure{
		\includegraphics[width=0.23\textwidth]{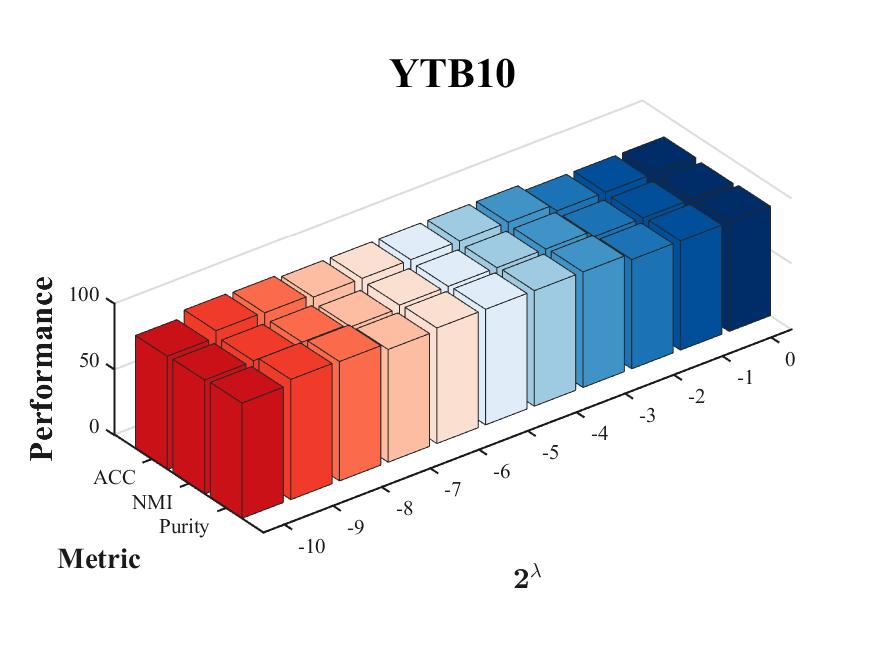}}
	%\subfigure{
	%	\includegraphics[width=0.3\textwidth]{pdf/olympics_iter.pdf}}
	
\centering
	\caption{The sensitivity of CCMVC-FSF with the variation of $\lambda$ on 3Sources, Mfeat, Reuters, and YTB10.}
\label{fig_lam}
\end{figure*}
Nevertheless, the codes of these compared algorithms are publicly available online, and we directly use them without changes. For methods with hyper-parameters, we follow the same strategies in the corresponding papers to tune them and report the best results. In our paper, the hyper-parameter $\lambda$ ranges from $2 .^{\wedge}[-10,-9, \cdots, 0]$. We evaluate the performance of algorithms via three widely used metrics, i.e., accuracy (ACC), normalized mutual information (NMI), and purity. To eliminate the randomness of k-means, each algorithm performs k-means 50 times and reports the average. Our experiments are implemented on a desktop computer with an Intel(R) Core(TM) i9-10850K CPU and 96 GB RAM.

% \usepackage[normalem]{ulem}

\subsection{Experimental Results}
To demonstrate the effectiveness of CCMVC-FSF, we conduct experiments to report the clustering results of CCMVC-FSF on six widely used benchmark datasets with the compared algorithms, and the results are shown in Table \ref{cmp_result}. From the table, it is observed that: 
\begin{itemize}
    \item CCMVC-FSF shows a clear advantage over the compared baselines. For instance, it consistently outperforms the second-best method over 4.47\%, 0.80\%, 5.36\%, 0.40\%, 2.28\%, and 6.09\% on ACC. The results on other metrics are also promising. The excellent clustering performance exhibits the effectiveness of CCMVC-FSF.
    \item Compared with graph-based continual multi-view clustering methods such as IMSC and SCGL, the late-fusion based MVC like CMVC and CCMVC-FSF achieves better clustering performance. Meanwhile, the space complexity is less, indicating that late fusion-based methods might be more qualified for continual multi-view data, and it inspires future research.
    \item Our proposed method exceeds CMVC on most datasets. Given that the main difference between them is that CCMVC-FSF stores filtered structural information of previous views to guide the partition matrix generation and CMVC does not, we can conclude that in continual multi-view learning, it is significant to store filtered prior structural information to guide the clustering process of the newly collected view.
    
\end{itemize}
\begin{table*}[]
\centering
	\caption{The effect of $r$ on our method of clustering performance on six datasets in terms of ACC.}
	\label{r_result}
\begin{tabular}{ccccccc}
\toprule
Datasets    & 3Sources & Mfeat & Citeseer & SUNRGBD & Reuters & YTB10 \\ \hline
$r=n^{1/4}$ & 66.86    & 93.20  & 47.89    & 19.36   & 50.23   & 88.17 \\
$r=n^{1/3}$ & 68.26    & 93.35 & 48.76    & 19.86   & 50.78   & 90.26 \\
$r=n^{1/2}$ & 69.23    & 94.95 & 49.37    & 19.94   & 51.65   & 91.44 \\
$r=n^{3/4}$ & 69.41    & 94.95 & 49.14    & 19.93   & 51.93   & 91.39 \\
$r=n$       & 70.01    & 95.00 & 49.83    & 19.99   & 52.03   & 92.00 \\ \bottomrule
\end{tabular}
\end{table*}


\begin{table*}[]
\centering
	\caption{The running time comparison of our method with different values of $r$.}
	\label{r_time}
\begin{tabular}{ccccccc}
\toprule
Datasets    & 3Sources & Mfeat & Citeseer & SUNRGBD & Reuters & YTB10  \\ \hline
$r=n^{1/4}$ & 0.07     & 6.42  & 20.10    & 17.98   & 134.10  & 62.61  \\
$r=n^{1/3}$ & 0.08     & 7.65  & 25.38    & 29.30   & 239.25  & 73.69  \\
$r=n^{1/2}$ & 0.09     & 10.10 & 42.09    & 43.39   & 346.96  & 81.32  \\
$r=n^{3/4}$ & 0.12     & 14.45 & 42.49    & 87.34   & 478.47  & 108.26 \\
$r=n$       & 0.27     & 26.14 & 68.83    & 196.54  & 767.06  & 160.49 
 \\ \bottomrule
\end{tabular}
\end{table*}
\begin{table*}[]
\centering
	\caption{The ablation study of our proposed method with five compared methods on six benchmark datasets in terms of clustering accuracy (ACC).}
	\label{ablation_result}

\begin{tabular}{ccccccc} \toprule
Datasets  & 3Sources & Mfeat & Citeseer & SUNRGBD & Reuters & YTB10 \\ \hline
IMVC      & 60.95    & 83.10 & 41.64    & 19.11   & 50.18   & 87.91 \\
CMVC-G    & 65.09    & 92.55 & 41.73    & 19.12   & 49.39   & 87.39 \\
CCMVC-S   & 63.86    & 83.05 & 46.28    & 19.20   & 50.21   & 86.74 \\
CCMVC-K   & 65.68    & 89.70 & 47.89    & 19.19   & 49.39   & 88.51 \\
CCMVC-RS  & 60.35    & 81.35 & 47.70    & 18.76   & 48.01   & 89.46 \\
CCMVC-FSF & 69.23    & 94.95 & 49.37    & 19.94   & 51.65   & 91.44 \\ \bottomrule
\end{tabular}

\end{table*}
\subsection{Convergence and Evolution}
As analyzed above, the alternating optimization of our proposed method is convergent theoretically. We plot the objective values varying with iterations on four datasets in Fig. \ref{fig_obj} to verify it experimentally. From the figure, we can conclude that the objective value increases monotonously and converges in less than 20 iterations, validating the convergence analysis in Section \ref{convergence_section} experimentally. Also, to investigate the quality of the consensus matrix varies with iterations, we conduct experiments to visualize the similarity matrix computed by $\mathbf{H}^*_{t}{\mathbf{H}^*_{t}}^{\top}$ on Mfeat and plot it in Fig. \ref{fig_matrix_qua}. It is obtained that the block diagonal structure of $\mathbf{H}^*_{t}{\mathbf{H}^*_{t}}^{\top}$ is enhanced with iterations, which benefits the subsequent clustering process.

\subsection{Parameter Sensitivity}
We introduce a regularization term to trade off partition matrix fusion and contrastive loss. To investigate the effect of the parameter on our model, we conduct experiments, and the results are shown in Fig. \ref{fig_lam}. From the figure, it is observed that the clustering performance of our proposed method remains stable with a wide range of $\lambda$.

\subsection{The Effect of \texorpdfstring{${r}$}-}
To reduce the high complexity of the filtered structural information extraction, we propose a clustering then sample strategy, and the suitable value of the sampling number $r$ is given by Remark \ref{r_value}. In this section, we conduct experiments to investigate the value of $r$ on our model. The clustering performance of our method with different values of $r$ is reported in Table \ref{r_result}, and the running time is recorded in Table \ref{r_time}. It is worth mentioning that when setting $r=n$, the method will search the positive/negative data from the whole data cloud. From the figure, we obtain that the value of $r$ has a great impact on the model. In most cases, larger $r$ makes the clustering performance better but results in more time consumption. Meanwhile, when the value of $r$ reaches a degree, the increase in clustering performance is unworthy. From the tables, we provide a more solid understanding of the algorithm properties as well as justification for our choices.

\begin{figure*}[htbp]
	\centering

	% \subfigure{
	% 	\includegraphics[width=0.23\textwidth]{pdf/ar10p_CFP.pdf}}
% \hspace{-0.2cm}
	\subfigure{
		\includegraphics[width=0.23\textwidth]{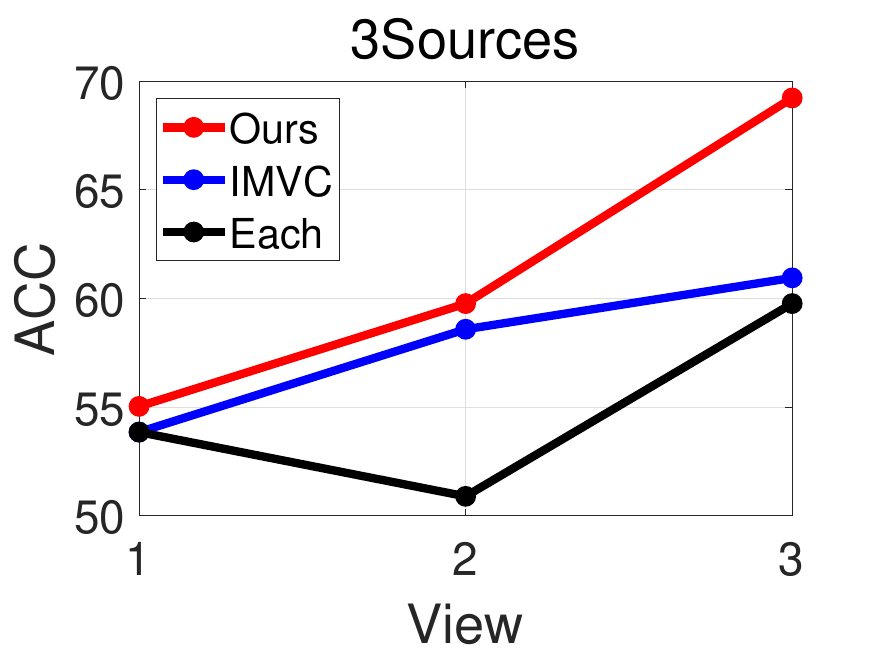}}
% \hspace{-0.2cm}
% 	\subfigure{
% 		\includegraphics[width=0.3\textwidth]{pdf/mfea.pdf}}
% % \hspace{-0.2cm}	
% \subfigure{
% 		\includegraphics[width=0.3\textwidth]{pdf/awa_iter.pdf}}
% \hspace{-0.2cm}
	\subfigure{
		\includegraphics[width=0.23\textwidth]{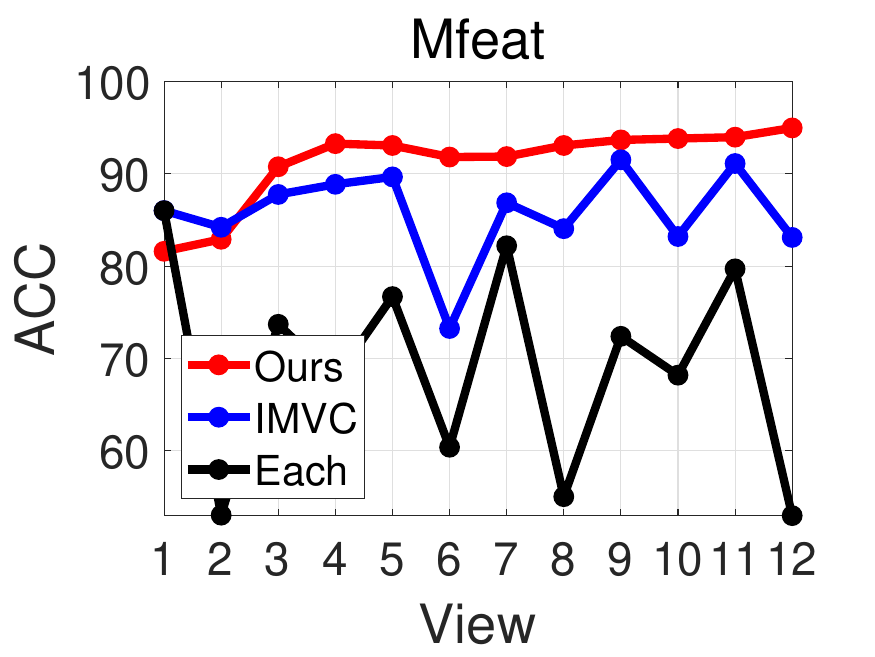}}
% \hspace{-0.2cm}
	\subfigure{
		\includegraphics[width=0.23\textwidth]{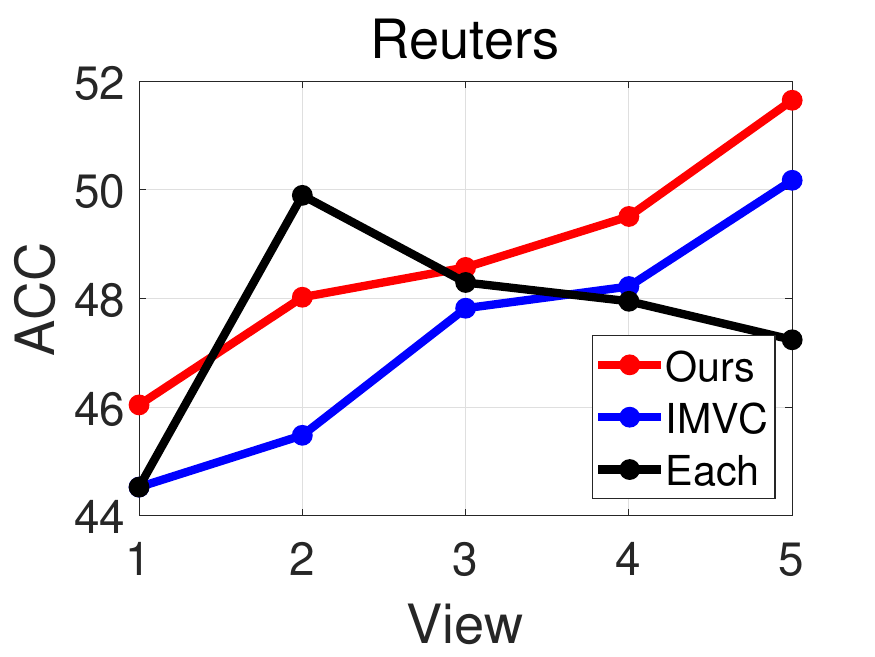}}
  \subfigure{
		\includegraphics[width=0.23\textwidth]{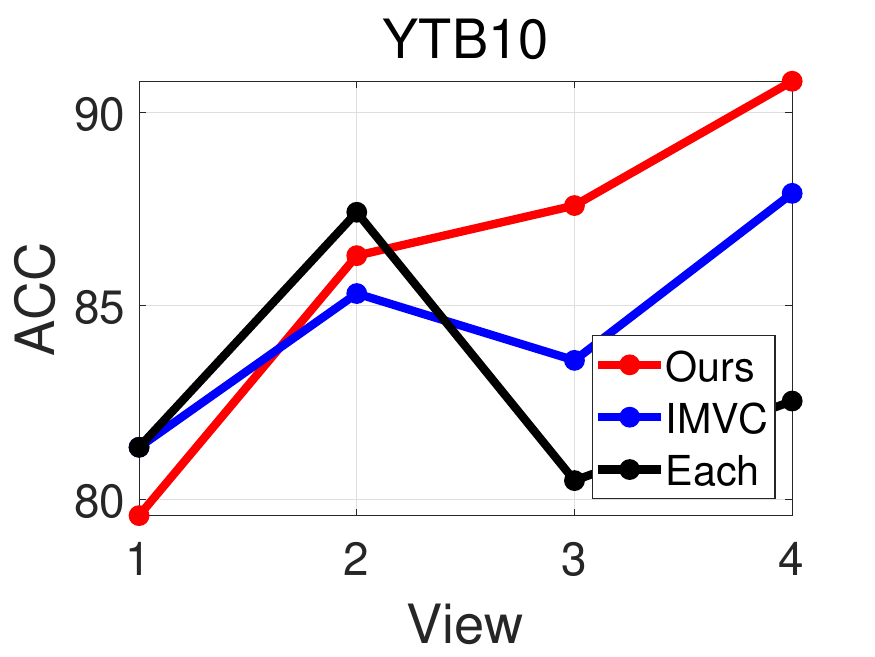}}
% \hspace{-0.2cm}	

	%\subfigure{
	%	\includegraphics[width=0.3\textwidth]{pdf/olympics_iter.pdf}}
	
\centering
	\caption{The effect on CFP with views collected and fused in order on 3Sources, Mfeat, Reuters, and YTB10 in terms of ACC, respectively.}
\label{fig_CFP}
\end{figure*}
\begin{figure*}[htbp]
	\centering

	% \subfigure{
	% 	\includegraphics[width=0.23\textwidth]{pdf/ar10p_cmp_CFP.pdf}}
% \hspace{-0.2cm}
	\subfigure{
		\includegraphics[width=0.23\textwidth]{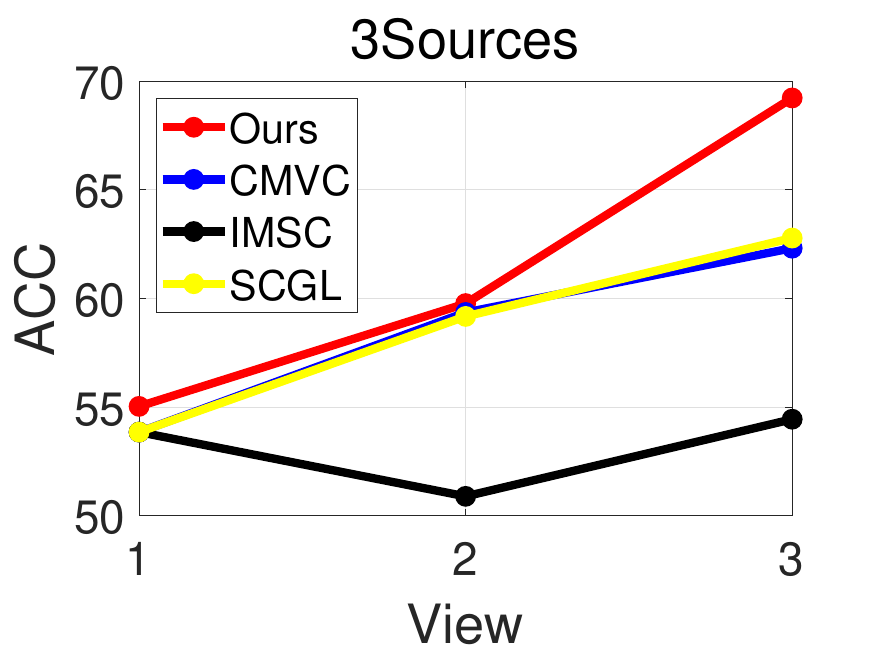}}
% \hspace{-0.2cm}
% 	\subfigure{
% 		\includegraphics[width=0.3\textwidth]{pdf/mfea.pdf}}
% % \hspace{-0.2cm}	
% \subfigure{
% 		\includegraphics[width=0.3\textwidth]{pdf/awa_iter.pdf}}
% \hspace{-0.2cm}
	\subfigure{
		\includegraphics[width=0.23\textwidth]{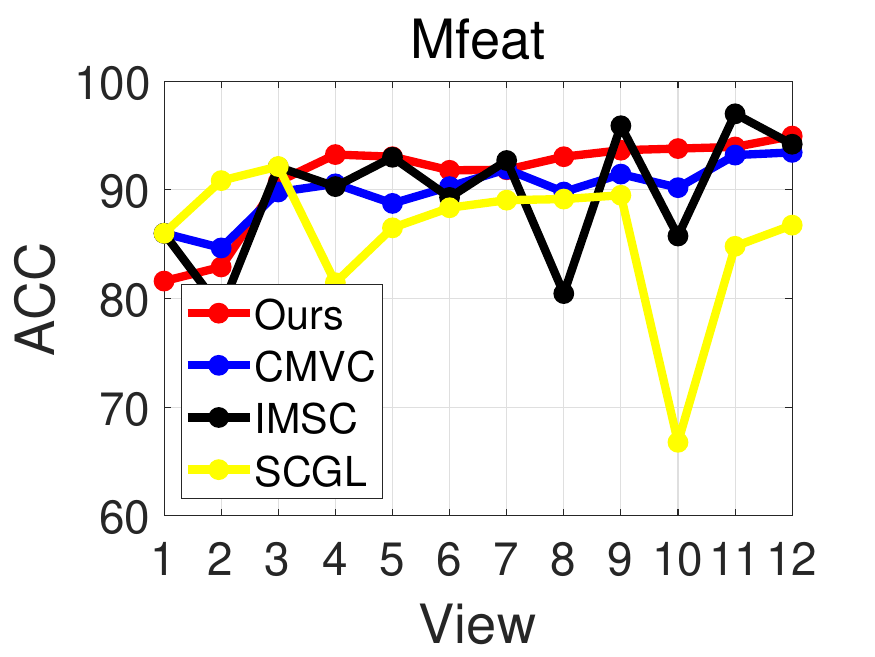}}
% \hspace{-0.2cm}
	\subfigure{
		\includegraphics[width=0.23\textwidth]{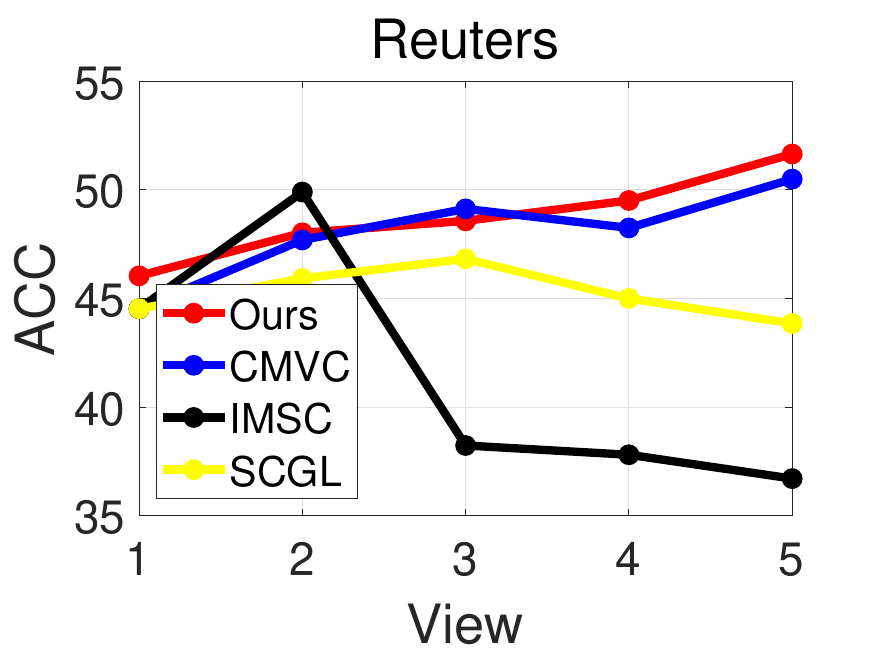}}
    \subfigure{
		\includegraphics[width=0.23\textwidth]{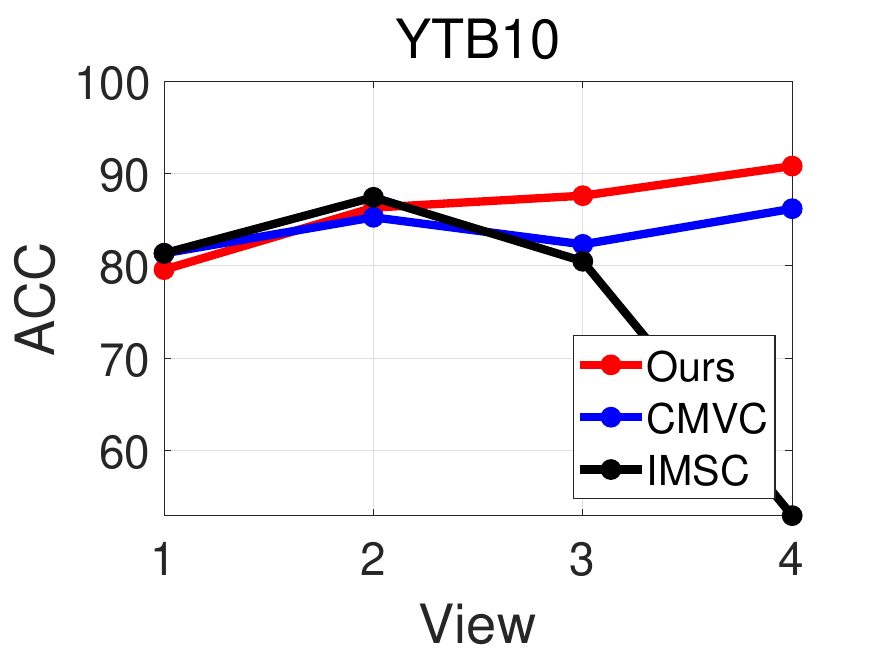}}
% \hspace{-0.2cm}	

	%\subfigure{
	%	\includegraphics[width=0.3\textwidth]{pdf/olympics_iter.pdf}}
	
\centering
	\caption{The effect on CFP with views collected and fused in order on four datasets of our method and three continual multi-view clustering methods.}
\label{fig_cmp_CFP}
\end{figure*}
 \begin{figure*}[]
	\centering

	\subfigure{
		\includegraphics[width=0.9\textwidth]{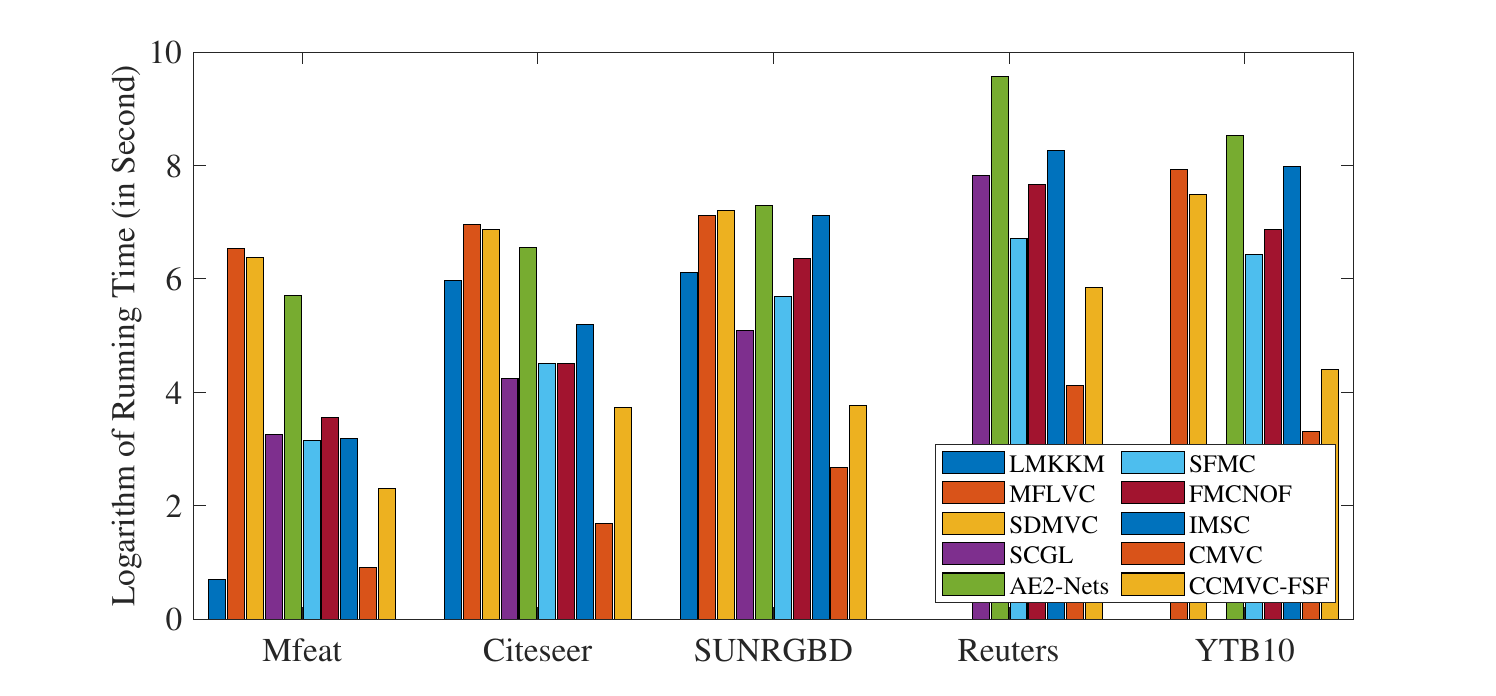}}

\centering
	\caption{Running time comparison of compared methods on datasets of size more than 1000.}\label{time_fig}
\end{figure*}
\subsection{Ablation Studies}
In this section, we study the effectiveness of the critical components in CCMVC-FSF. We develop a fixed data buffer to store filtered structural information and utilize it to conduct contrastive learning to guide the generation of the consensus matrix when a new view arrives. Meanwhile, the positive/negative pairs are generated by an efficient clustering then sample method in each view, and we discuss why we do not conduct $k$-means and directly assign samples in the same cluster as positive pairwise. To investigate its effectiveness, we design five methods to compare and report the results in Table \ref{ablation_result}, where IMVC is a particular case of CCMVC-FSF by setting $\lambda=0$ in Eq. \eqref{final_CCMVC}, CMVC-G stores unfiltered graph information and use it to guide the generation of a consensus matrix, CCMVC-S utilizes the sample itself as positive samples and others as negative samples, likewise existing contrastive learning methods, and CCMVC-K assigns samples in the same cluster as positive pairwise via a $k$-means algorithm, CCMVC-RS conducts random selection to get positive/negative pairs. From the table, it is illustrated that: 1) Compared with IMVC, our method gains better clustering performance, which indicates that in continual multi-view clustering, utilizing previous view information to conduct clustering is substantial. 2) Our method outperforms CCMVC-G, showing that the filtered structural information is superior to unfiltered information in continual multi-view clustering. The reason might be that the correlations among all samples are redundant and sensitive to views with poor quality. 3) CCMVC-FSF is superior to CCMVC-S, CCMVC-K, and CCMVC-RS, showing that our proposed positive/negative pairwise settings are more rational than existing settings.

\subsection{The Effect on CFP}
We propose a first study on the catastrophic forgetting problem (CFP) in multi-view clustering and utilize filtered structural information to cope with it. To investigate the effect of CCMVC-FSF on CFP, we conduct experiments with views collected with time and plot the results of views integrated one by one in Fig. \ref{fig_CFP} and Fig. \ref{fig_cmp_CFP}. In Fig. \ref{fig_CFP},  ’Each' denotes the clustering performance on each view without other information. From the figure, we obtain that the attempt to utilize previous information to guide the clustering process is significant in continual multi-view clustering. For instance, IMVC is sensitive to the view with poor performance, while CCMVC-FSF is more robust. Meanwhile, when the number of previous views is abundant, our proposed method remains a promising performance even though a view arrives with poor clustering quality, such as on Mfeat. Precisely, the clustering quality of the sixth view is poor, but the results of our method do not decrease significantly, while the compared ones drop obviously. In Fig. \ref{fig_cmp_CFP}, we compare the effect on CFP of our method with existing continual multi-view clustering methods. It is obtained that existing methods undergo a CFP problem and are sensitive to the newly collected view, while our proposed method alleviates it. For instance, on Reuters, the clustering performance of the compared three methods undergoes prominent fluctuations, while CCMVC-FSF is more robust.
\subsection{Running Time}
We conduct experiments to record the running time of our proposed algorithm and the compared ones, and the results are shown in Figure \ref{time_fig}. It is worth noting that the running time of the continual multi-view clustering methods starts with the first views arriving and ends with the fusion of the last view. From the figure, we observe that CCMVC-FSF demonstrates a comparable time efficiency among the state-of-the-art multi-view clustering methods. Also, in scenarios where views are collected with time, traditional MVC methods cost more time because they need to integrate all the views repeatedly when a new view arrives, privacy issues might prevent the historical data from being stored, while continual MVC overcomes this problem. Although CMVC costs less time, our proposed method outperforms it on all datasets in terms of ACC. Also, CMVC undergoes the CFP problem and is unsuitable for practical applications. Therefore, the time spent on CCMVC-FSF is worthwhile.

\section{Conclusion}
In this paper, we propose the first study on the catastrophic forgetting problem in multi-view clustering. A novel method termed Contrastive Continual Multi-view Clustering with Filtered Structural Fusion (CCMVC-FSF) is provided to solve the problem. We store the filtered structural information of previous views in a fixed buffer by a clustering then sample method, and the suitable sampling number is analyzed. When a new view arrives, the filtered structural information will guide its clustering process as a contrastive term. The resultant problem is solved via an alternating optimization strategy with proven convergence. The theoretical analysis between CCMVC-FSF with semi-supervised learning and knowledge distillation is provided. Extensive experiments exhibit its superiority. In the future, we intend to extract knowledge of previous views under a deep learning framework.

% or


\ifCLASSOPTIONcompsoc
  \section*{Acknowledgments}
\else
  \section*{Acknowledgment}
\fi

This work is supported by the National Natural Science Foundation of China (project no. 61972056, 62071157, 62276271, 62306324, 62325604, 62376039, and 62376279).

\ifCLASSOPTIONcaptionsoff
  \newpage
\fi



%\bibliographystyle{IEEEtran}

\begin{IEEEbiography}[{\includegraphics[width=1in,height=1.10in,clip,keepaspectratio]{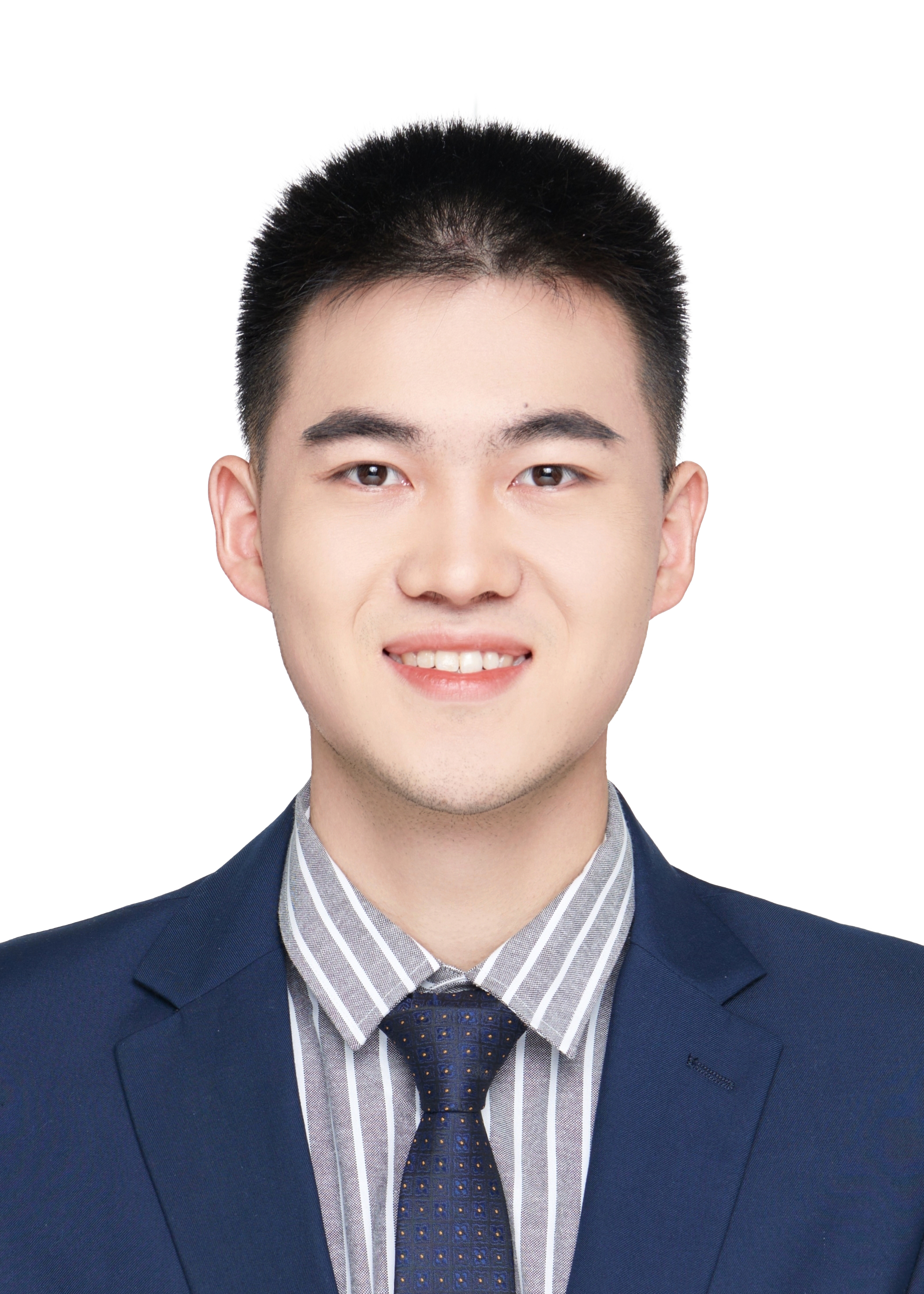}}]{Xinhang Wan} received the B.E degree in Computer Science and Technology from Northeastern University, Shenyang, China, in 2021. He is currently working toward the Master’s degree with the National University of Defense Technology (NUDT), Changsha, China. He has published papers in journals and conferences such as IEEE T-NNLS, ACMMM, AAAI, etc. His current research interests include multi-view learning, continual clustering, and active learning.
\end{IEEEbiography}

% \begin{IEEEbiography}[{\includegraphics[width=1in,height=1.05in,clip,keepaspectratio]{AuthorFigures/QingLiao.jpg}}]{Qing Liao} received her Ph.D. degree in computer science and engineering in 2016 supervised by Prof. Qian Zhang from the Department of Computer Science and Engineering of the Hong Kong University of Science and Technology. She is currently a professor with School of Computer Science and Technology, Harbin Institute of Technology (Shenzhen), China. Her research interests include artificial intelligence and data mining.
% \end{IEEEbiography}
\begin{IEEEbiography}[{\includegraphics[width=1in,height=1.10in,clip,keepaspectratio]{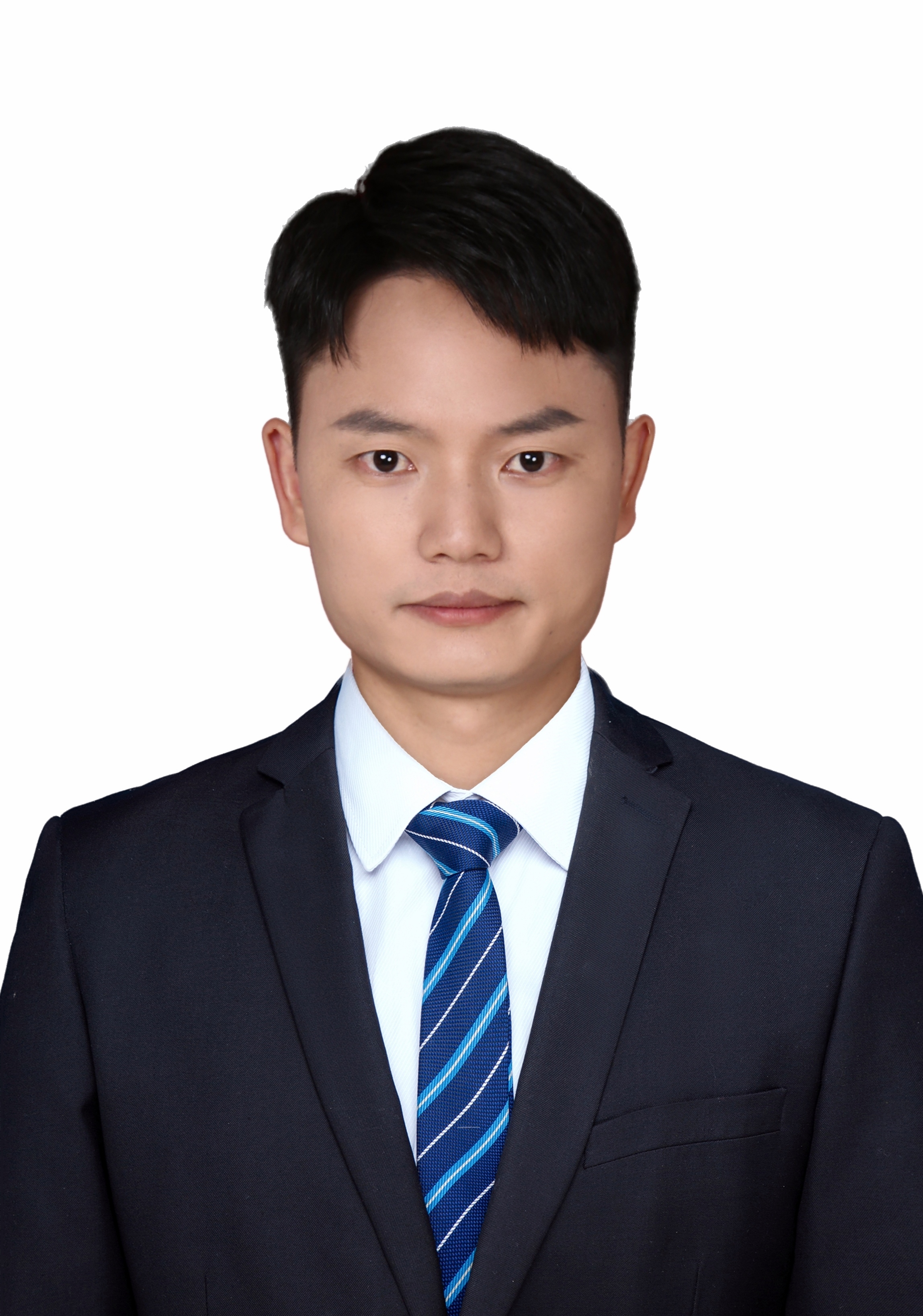}}]
{Jiyuan Liu} received his PhD from National University of Defense Technology (NUDT), China, in 2022. He is now a lecturer with the College of Systems Engineering, NUDT. His current research interests include multi-view clustering, federated learning and anomaly detection. Dr. Liu has published papers in journals and conferences such as IEEE TPAMI, IEEE TKDE, IEEE TNNLS, ICML, NeurIPS, CVPR, ICCV, etc. He serves as program committee member and reviewer on IEEE TPAMI, IEEE TKDE, IEEE TNNLS, ICML, NeurIPS, CVPR, ICCV, etc. More information can be found at \url{https://liujiyuan13.github.io/}.
\end{IEEEbiography}

\begin{IEEEbiography}
[{\includegraphics[width=0.8in,clip,keepaspectratio]{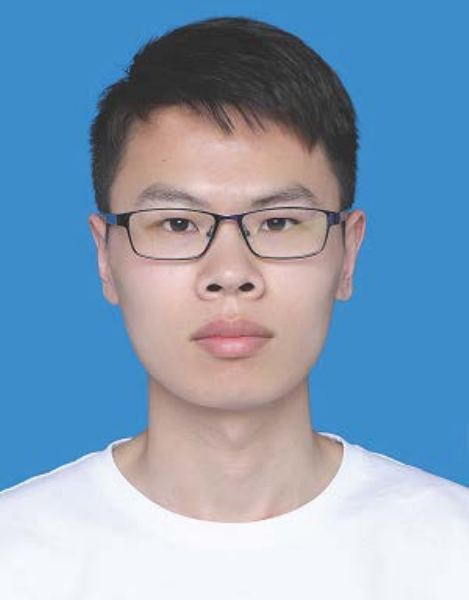}}]
{Hao Yu} is presently pursuing a Ph.D. degree at the National University of Defense Technology (NUDT).
He earned a B.Eng degree in computer science from Inner Mongolia University, Hohhot, China, in 2019, and, subsequently, in 2022, obtained an MA.Sc in cyberspace science and technology from Beijing Institute of Technology, Beijing, China.
His current research focuses on AI security and Federated Learning.
He has authored several papers in top-level journals and conferences, such as IEEE TIFS, TDSC, and ACM MM, and served as a Reviewer for highly regarded journals, such as IEEE TIFS, IEEE TKDE, ACM TOIS, etc.
\end{IEEEbiography}

\begin{IEEEbiography}[{\includegraphics[width=1in,height=1.10in,clip,keepaspectratio]{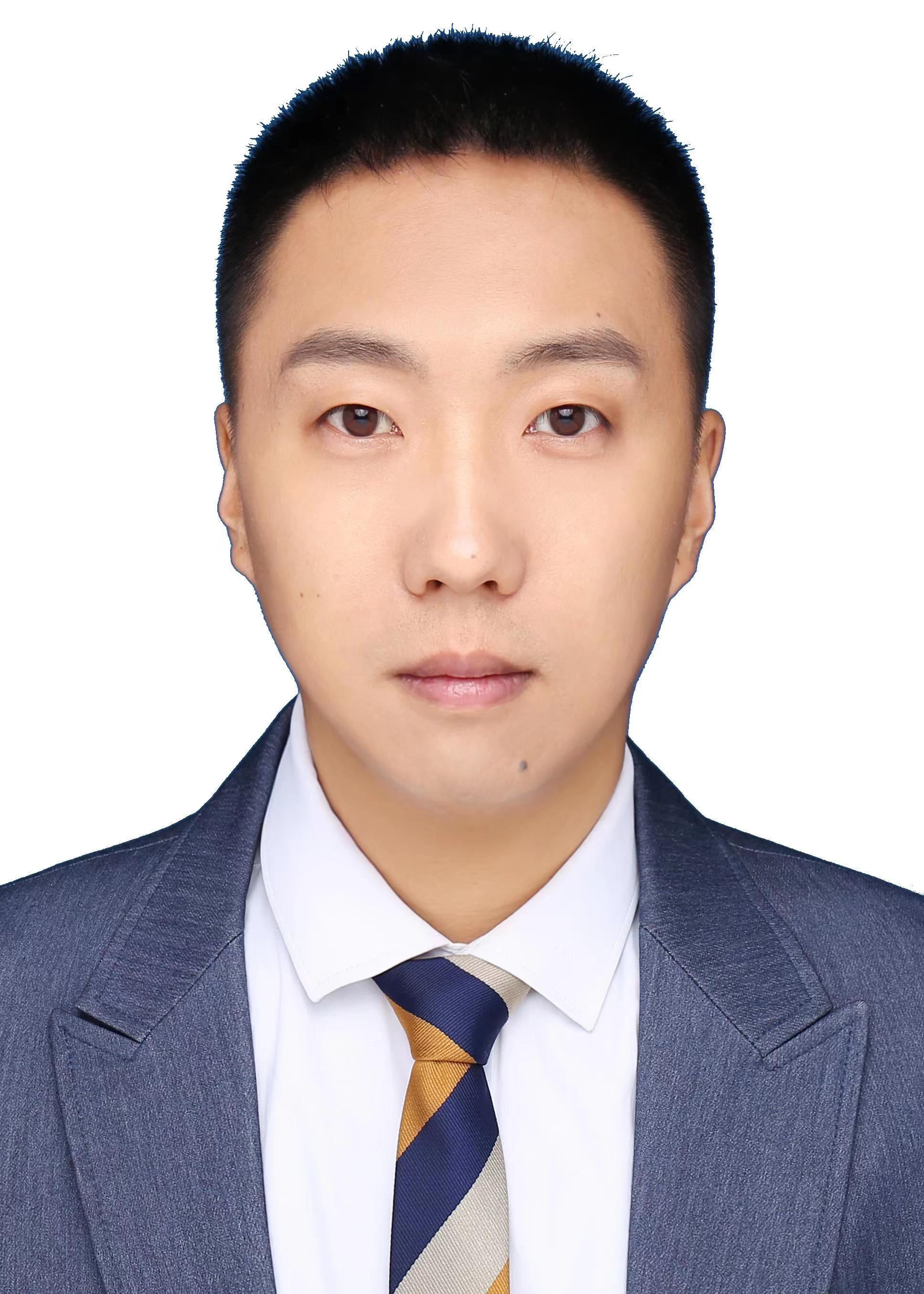}}]{Ao Li} received his PhD degree from Harbin Engineering University, China. He is now a professor at the School of Computer Science and Technology, Harbin University of Science and Technology, China. He also serves as a research assistant with Wright State University from 2017 to 2018. His main research interests include pattern recognition and machine learning. Dr. Li has published 30+ peer-reviewed papers, including those in IEEE TR, Information Sciences, Knowledge-based Systems, etc.
\end{IEEEbiography}

\begin{IEEEbiography}[{\includegraphics[width=1in,height=1.10in,clip,keepaspectratio]{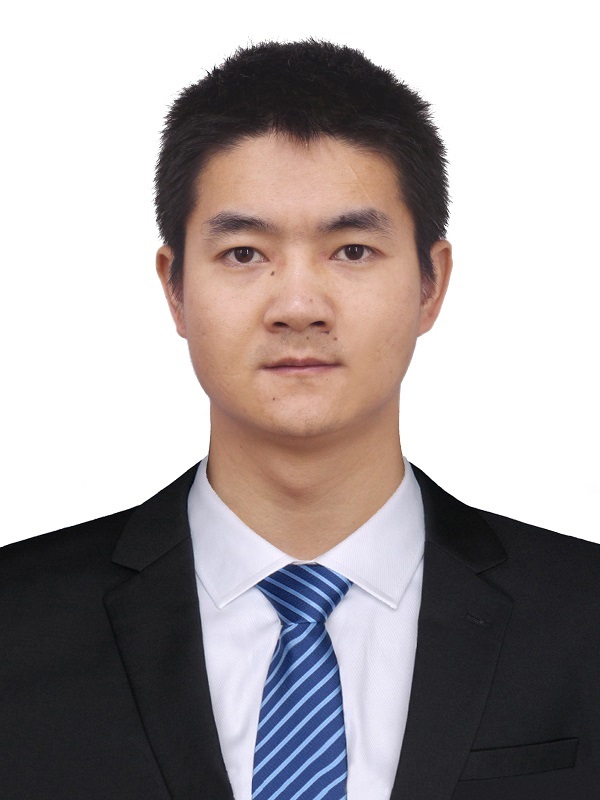}}]
{Xinwang Liu} received his PhD degree from National University of Defense Technology (NUDT), China. He is now Professor of School of Computer, NUDT. His current research interests include kernel learning and unsupervised feature learning. Dr. Liu has published 90+ peer-reviewed papers, including those in highly regarded journals and conferences such as IEEE T-PAMI, IEEE T-KDE, IEEE T-IP, IEEE T-NNLS, IEEE T-MM, IEEE T-IFS, ICML, NeurIPS, ICCV, CVPR, AAAI, IJCAI, etc. He serves as the associated editor of IEEE TNNLS/IEEE TCYB/Information Fusion Journal. More information can be found at \url{https://xinwangliu.github.io/}.
\end{IEEEbiography}

\begin{IEEEbiography}[{\includegraphics[width=1in,height=1.10in,clip,keepaspectratio]{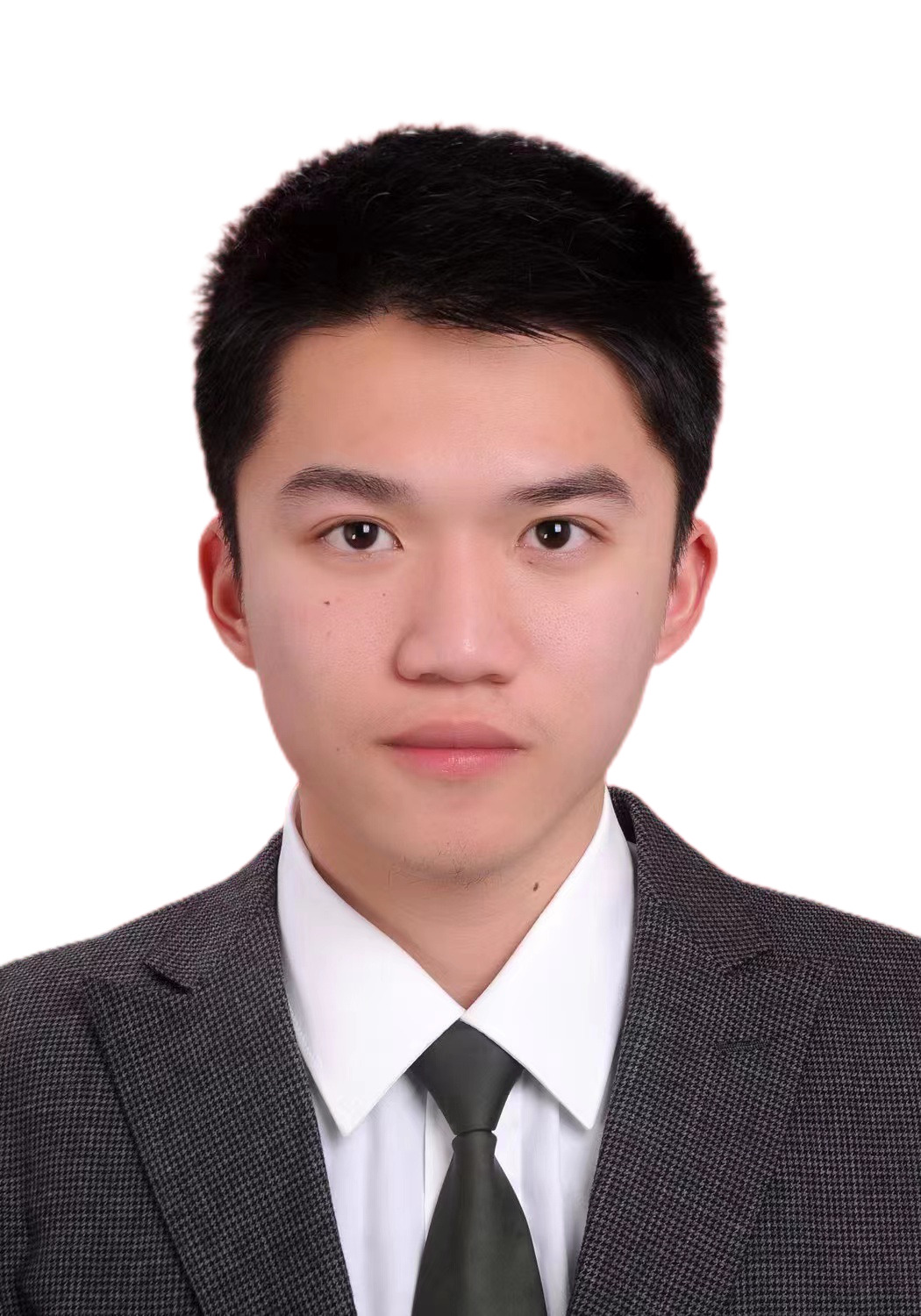}}]{Ke Liang} is currently pursuing a Ph.D. degree at the National University of Defense Technology (NUDT). Before joining NUDT, he got his BSc degree at Beihang University (BUAA) and received his MSC degree from the Pennsylvania State University (PSU). His current research interests include graph representation learning, multi-modal representation learning, and medical image processing.
\end{IEEEbiography}

\begin{IEEEbiography}[{\includegraphics[width=1in,height=1.10in,clip,keepaspectratio]{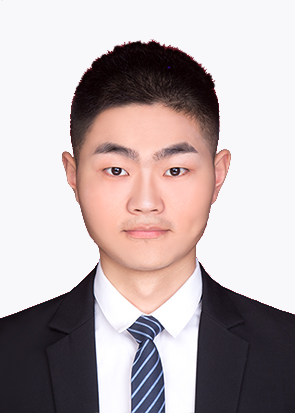}}]
{Zhibin Dong} is currently pursuing the Ph.D. degree with the National University of Defense Technology (NUDT), Changsha, China. He has published several papers and served as a Program Committee (PC) member or a reviewer for top conferences, such as IEEE Conference on Computer Vision and Pattern Recognition (CVPR), IEEE International Conference on Computer Vision(ICCV), Association for Computing Machinery’s Multimedia Conference (ACM MM), Association for the Advancement of Artificial Intelligence Conference (AAAI), and the IEEE TRANSACTIONS ON NEURAL NETWORKS AND LEARNING SYSTEMS (TNNLS). His current research interests include graph representation learning, deep unsupervised learning, and multiview clustering.
\end{IEEEbiography}

\begin{IEEEbiography}[{\includegraphics[width=1in,height=1.10in,clip,keepaspectratio]{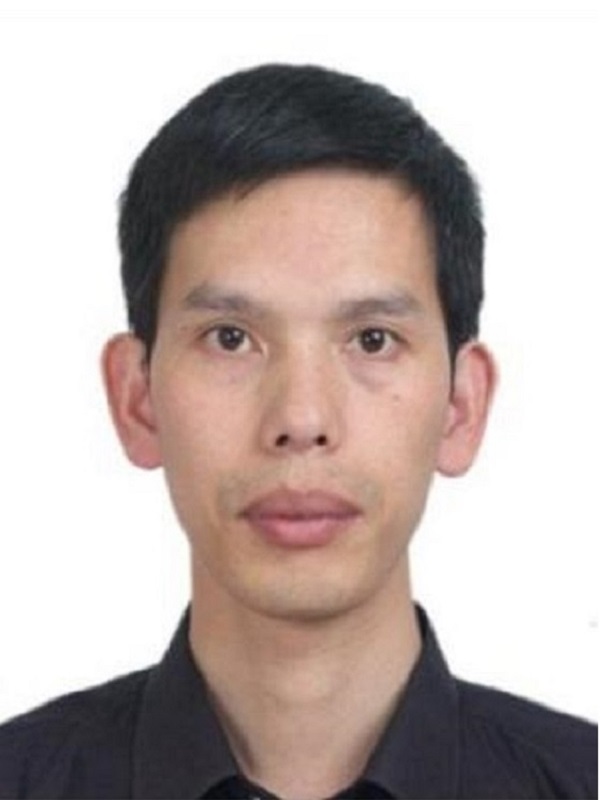}}]{En Zhu} received the Ph.D. degree from the National University of Defense Technology (NUDT), China. He is currently a Professor with the School of Computer Science, NUDT. He has published more than 60 peer-reviewed papers, including IEEE TRANSACTIONS ON CIRCUITS AND SYSTEMS FOR VIDEO TECHNOLOGY (TCSVT), IEEE TRANSACTIONS ON NEURAL NETWORKS AND LEARNING SYSTEMS (TNNLS), PR, AAAI, and IJCAI. His main research interests include pattern recognition, image processing, machine vision, and machine learning. He was awarded the China National Excellence Doctoral Dissertation.
\end{IEEEbiography}


%\vfill


\end{document}